\pdfoutput=1

\documentclass[11pt]{article}

\usepackage{EMNLP2023}

\usepackage{times}
\usepackage{latexsym}

\usepackage[T1]{fontenc}

\usepackage[utf8]{inputenc}

\usepackage{microtype}

\usepackage{inconsolata}
\usepackage{amsmath}
\usepackage{multirow}

\usepackage{xcolor}

\usepackage{algorithm}
\usepackage{algorithmic}
\usepackage{subcaption}
\usepackage{cleveref}
\crefname{equation}{Eq.}{Eq.}
\crefname{section}{Section}{Sections}
\crefname{subsection}{Section}{Sections}
\crefname{subsubsection}{Section}{Sections}
\crefname{figure}{Figure}{Figures}
\crefname{table}{Table}{Tables}
\crefname{subfigure}{Figure}{Figures}
\crefname{algocf}{Algorithm}{Algorithms}

\usepackage{newfloat}
\usepackage{listings}
\usepackage{booktabs}
\usepackage{graphicx}

\newcommand{\ours}{\textsc{MedEval}}
\title{M4:  Multi-Task Medical Evaluation Benchmark\\ with Expert Multi-Grained Labels across Multiple Domains}
\title{MediBench: A Comprehensive Multi-level and Multi-task Benchmark with Curated Medical Records }

\title{Generalize to the Niche? Multi-level and Multi-task Curated Medical Benchmark for LLM Generalizability  
}
\title{Exploring LLM Generalizability: A Curated Multi-task Medical Benchmark for Niche Domains}
\title{Assessing the Generalizability of Large Language Models through Benchmarking on a Niche and Curated Medical Dataset}

\title{M4: A Multi-Level, Multi-Task, and Multi-Domain Medical \\ Benchmark for Language Model Evaluation
}
\title{\ours: A Multi-Level, Multi-Task, and Multi-Domain Medical \\ Benchmark for Language Model Evaluation
}

\author{Zexue He\textsuperscript{\rm $\diamondsuit$},
Yu Wang\textsuperscript{\rm $\diamondsuit$},  
An Yan\textsuperscript{\rm $\diamondsuit$},  
Yao Liu\textsuperscript{\rm $\diamondsuit$}, 
Eric Y. Chang\textsuperscript{\rm $\diamondsuit$,$\clubsuit$}, 
\\\textbf{Amilcare Gentili\textsuperscript{\rm $\diamondsuit$,$\clubsuit$}, 
Julian McAuley\textsuperscript{\rm $\diamondsuit$}, 
Chun-Nan Hsu\textsuperscript{\rm $\diamondsuit$,$\clubsuit$}}\\
\textsuperscript{\rm $\diamondsuit$}University of California, San Diego, La Jolla, CA, United States \\
\textsuperscript{\rm $\clubsuit$}Veterans Affairs San Diego Healthcare System, San Diego, CA, United States\\
\texttt{\{zehe, yuw164, ayan, yal004, e8chang\}@ucsd.edu} \\
\texttt{\{agentili, jmcauley, chunnan\}@ucsd.edu} \\
}
\begin{document}
\maketitle
\begin{abstract}
Curated datasets for healthcare are often limited due to the need of human annotations from experts. In this paper, we present \ours, a multi-level, multi-task, and multi-domain medical benchmark to facilitate the development of language models for healthcare.
\ours ~is comprehensive and consists of data from several healthcare systems and spans 35 human body regions from 8 examination modalities.
With 22,779 collected sentences and 21,228 reports, we provide expert annotations at multiple levels, offering a granular potential usage of the data and supporting a wide range of tasks.
Moreover, we systematically evaluated 10 generic and domain-specific language models under zero-shot and finetuning settings, from domain-adapted baselines in healthcare to general-purposed state-of-the-art large language models (e.g., ChatGPT). 
Our evaluations reveal varying effectiveness of the two categories of language models across different tasks, from which we notice the importance of instruction tuning for few-shot usage of large language models. 
Our investigation paves the way toward benchmarking language models for healthcare and provides valuable insights into the strengths and limitations of adopting large language models in medical domains, informing their practical applications and future advancements\footnote{We will release the data set and 
source codes 
to facilitate the development and benchmarking of future healthcare language models
at \url{https://github.com/ZexueHe/MedEval} 
}.

\end{abstract}

\begin{figure}[th]
    \centering
    \begin{subfigure}[t]{\linewidth}
        \centering
        \includegraphics[width=\textwidth]{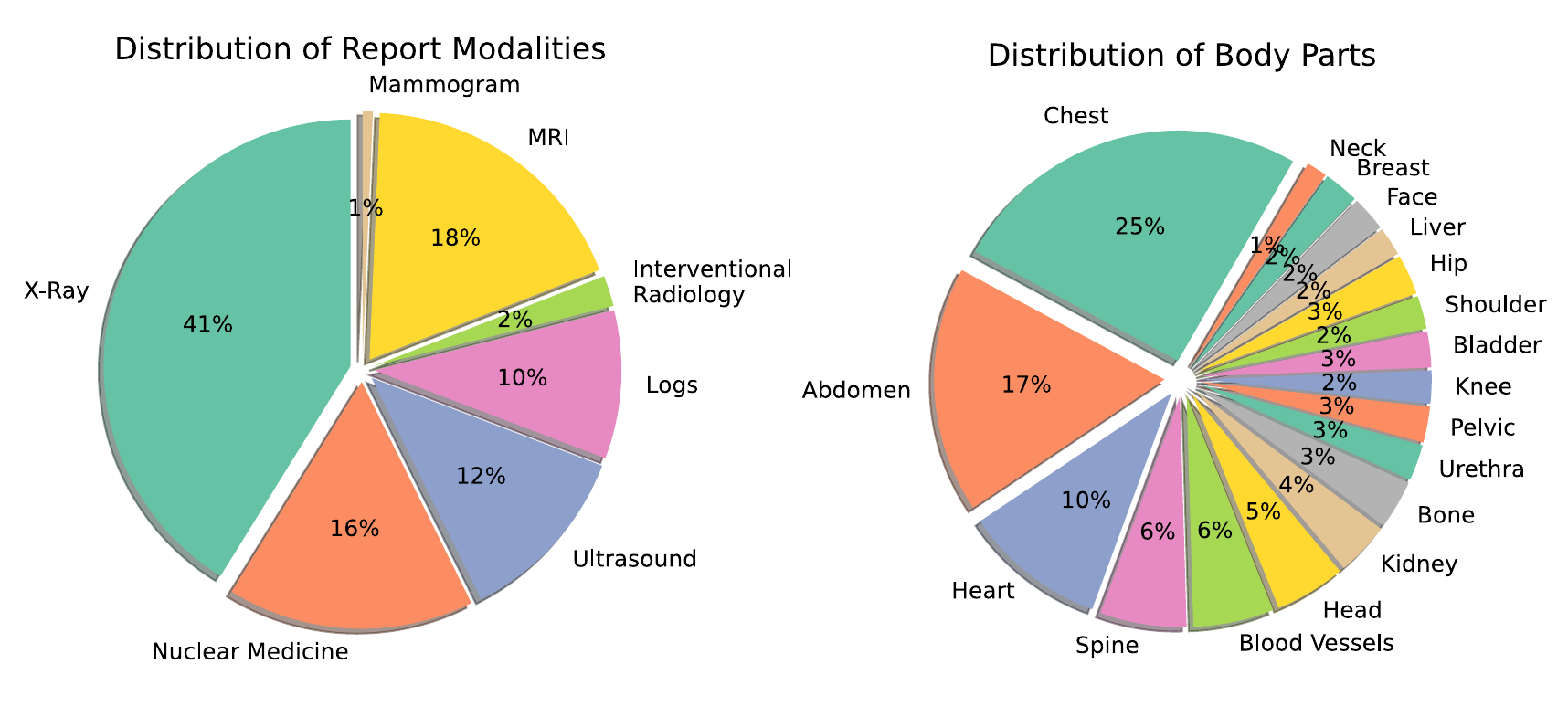}
        \caption{\small Distribution of modality and body parts  in \ours }
    \end{subfigure}%
    \hfill
    \begin{subfigure}[t]{\linewidth}
        \centering
        \includegraphics[width=\textwidth]{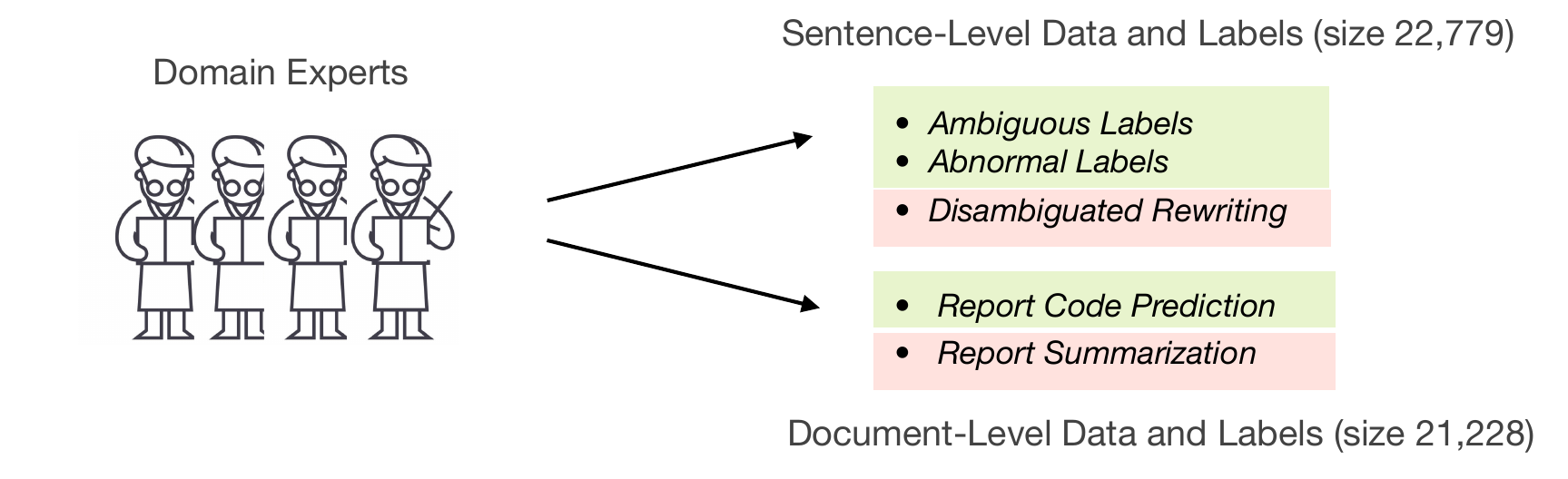}
        \caption{Multi-task expert labels at multi-granularity }
    \end{subfigure}
    \caption{A summary of the multi-level multi-task and multi-domain medical benchmark (\ours). Classification tasks are highlighted in green and generation tasks are highlighted in red.}
    \label{fig: M4}
\end{figure}
\section{Introduction}
Recent advanced language models, e.g., GPT-3, ChatGPT, and LLaMa \cite{llama},
are effective in various general tasks, suggesting their potential to healthcare use cases, such as alleviating the burden on human experts in decision-making and patient care. 
However, training, adapting, and evaluating these models requires high-quality domain-specific datasets, which are often challenging to obtain. Previous medical datasets have been collected from healthcare-related literature \cite{dernoncourt2017pubmed,gupta2021sumpubmed, jin2019pubmedqa, banarescu2013abstract} or web pages on the Internet \cite{mccreery2020effective, ammar2018construction}. While these datasets are large, they may lack quality with heterogeneous topics (e.g., scientific literature about nutrition may offer limited help in the decision-making process of an X-ray analysis). On the other hand, high-quality clinical data is typically obtained by annotating records from healthcare systems like MIMIC-CXR \cite{johnson2019mimic,yan2021weakly}. However, such data is either limited in size \cite{tsatsaronis2015overview}, 
or may only cover certain dominant systems and specific domains\footnote{We use \textit{domain} to describe data sets that are different in distribution, caused by the modality of the examination (e.g., X-ray, CT, Ultrasound, etc) and examined body parts (e.g., chest, abdomen, etc).} , such as chest X-rays~\citep{johnson2016mimic} or eye diseases ~\citep{otmakhova2022m3}. 
Other approaches involve automatically generating  medical corpus using templates~\citep{pampari2018emrqa, pappas2018bioread} or using language models~\citep{guo2023dr, tang2023does}, but they 
have been noted
to be limited in diversity, complexity, and quality \cite{gupta2021sumpubmed}. 

To tackle the aforementioned challenges and facilitate research in clinical NLP, we introduce \ours, a large-scale medical benchmark with multi-level curated labels for multiple tasks and multiple domains. \ours~comprises 22,779 sentence-level datapoints from radiology reports, including expert-crafted classification labels (e.g., abnormality identification labels) and ground truth for generation tasks (e.g., disambiguated rewritings). Additionally, we include 21,228 complete reports with expert-annotated medical codes for disease classification (e.g., for ankle radiology studies) and golden output for generation tasks (e.g., summarization of radiology reports). Besides the ability to support multi-tasks at different levels,  \ours's uniqueness also lies in its  diverse data coverage for different body parts (such as chest, foot, and ankle) and different modalities (X-rays, ultrasound, etc.), and the incorporated novel tasks/data that are collected from the U.S.
Department of Veterans Affairs (VA) health care system nationwide. 
To the best of our knowledge, \ours~represents the first expert-curated medical NLP benchmark that is both comprehensive and large-scale. \ours~ will be released to facilitate future research.

We further conduct a comprehensive evaluation of multiple state-of-the-art language model baselines, including domain-adapted PLMs followed by in-domain fine-tuning (e.g., fine-tuned BERT \cite{devlin2018bert}) and general-purposed LLMs utilized with few-shot in-context learning (e.g., ChatGPT). We evaluate their performance on sentence-level and document-level NLU and NLG tasks. We observe the effectiveness of both categories of models in different healthcare tasks, with surprisingly comparable performances from LLMs only using few-shot learning to domain-adapted PLMs in certain generation tasks. Our comprehensive evaluation indicates language models are strong candidates in medical tasks whose data is already seen/similar to their training data.
Our investigation provides  insights into the potentials and limitations of LLMs in healthcare domains, guiding the appropriate use of LLM-assisted healthcare decision-making systems in the future. Overall, our contributions are summarized as:
\begin{itemize}
    \item We propose a large-scale medical benchmark, namely \ours, with a broad coverage for various tasks and domains to facilitate future research in clinical NLP.
    \item We provide expert annotations for multiple tasks with multi-granularity, from sentence classification and rewriting, to report classification and summarization.
    \item We systematically evaluate various language models, and shed light on the strengths and weaknesses of these models for healthcare applications. 
\end{itemize}

\begin{figure*}[ht]
  \centering 
  \includegraphics[width=0.97\textwidth]{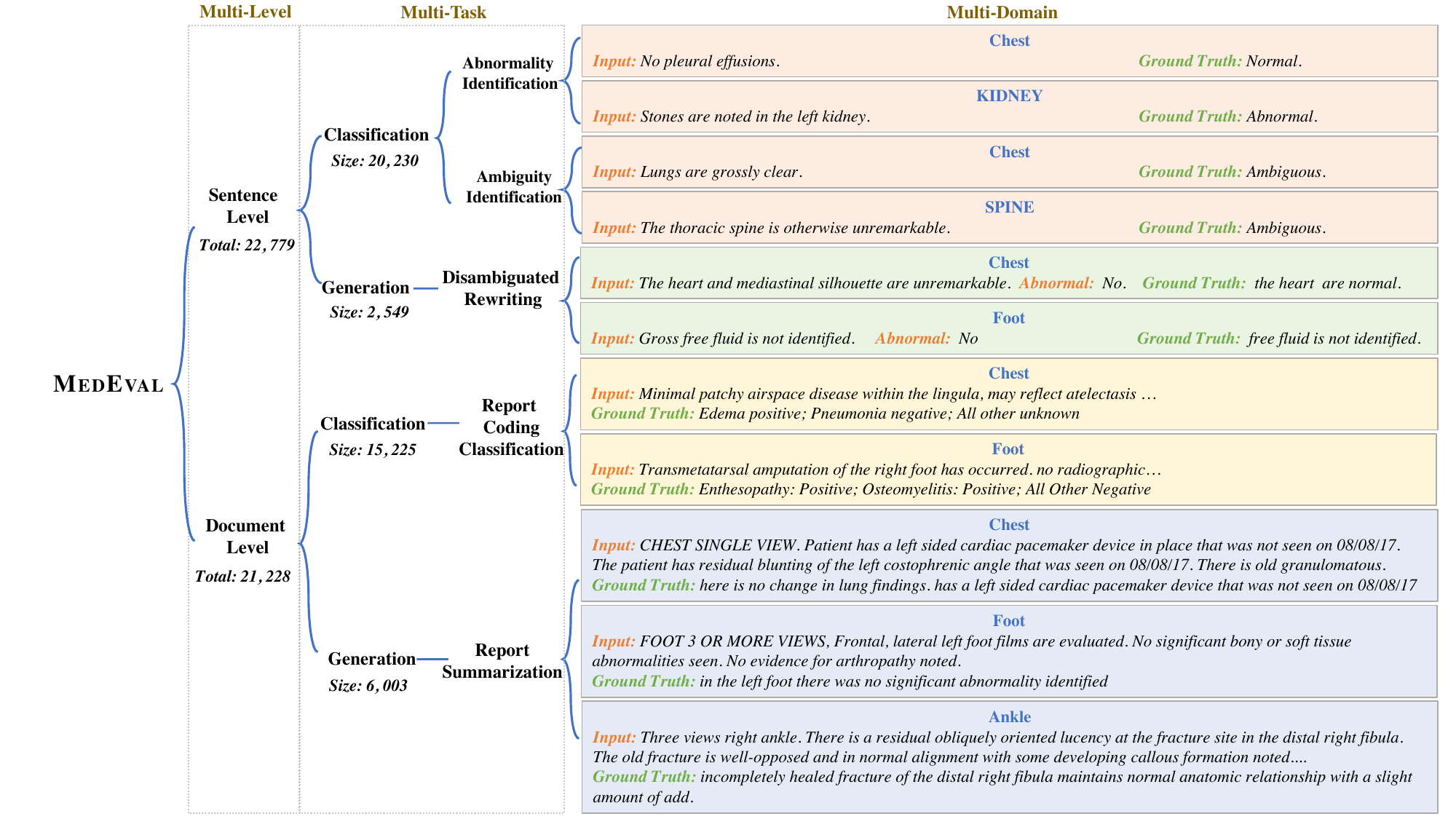}  
  \caption{\small Dataset composition of \ours. \ours~is a large-scale benchmark composed of 22,779 report sentences and 21,228 reports, covering multiple exam modalities on diverse body parts. 
  }
  \label{fig:dataset}
\end{figure*}
\section{Related Work}
\label{sec:related work}
\paragraph{Medical Benchmark} 


Existing medical benchmarks are typically collected from the following resources. First, data is crawled from public resources, such as biomedical journals and literature like PubMed Central (PMC) or Semantic Scholar \citet{ammar2018construction}, or healthcare-related web pages such as MQP \cite{mccreery2020effective} which was proposed by collecting COVID-19 FAQs from the Internet. Those data are heterogeneous and may lack relevance for assisting certain clinical purpose.  

On the other hand, high-quality benchmarks are extensively collected from real-world healthcare systems,
such as the  Medical Information Mart for Intensive Care databases (MIMIC \cite{johnson2016mimic, johnson2019mimic, johnson2023mimic}).
Several works have been proposed based on their databases. For instance, MIMIC-CXR \cite{johnson2019mimic} is a dataset consisting of pairs of radiology images and reports of chest X-ray exams. MIMIC PERform Dataset \cite{charlton2022detecting} comprises physiological signals related to critically-ill patients. Additionally, \citet{edin2023automated} collected document summary pairs labeled with diagnosis and procedure codes from \citet{johnson2023mimic}.  Though their collection may be one of the most comprehensive, there are many domains not included.  A more recent attempt was made to alleviate the incompleteness concern by introducing M3 \cite{otmakhova2022m3}, a multi-domain medical benchmark that incorporates multi-level expert annotations. 
By only considering studies on 
ophthalmology, M3 offers limited help in providing a fully comprehensive solution. 


To complement these existing efforts, we collect a large-scale dataset of real medical reports from another healthcare system that offers broader coverage including 35 human body regions from 8 examination modalities (e.g., X-ray, CT, etc.).
\paragraph{Language Models for Healthcare}
Large pre-trained language models are being widely adopted to solve healthcare tasks. One line of research involves adapting general language models to the biomedical domain through continuous training on domain-specific data and tasks. 
For instance, \citet{yan-etal-2021-weakly-supervised} enhanced BERT with contrastive learning for chest report generation. ClinicBERT \cite{huang2019clinicalbert} was proposed by continuously training BERT on clinic notes using masked language modeling, 
and \citet{yan2022radbert} developed RadBERT by continuously training BERT on a vast collection of radiology reports. Other adaptations of BERT, such as BioBERT \cite{lee2020biobert}, BlueBERT \cite{peng2019transfer}, SciBERT \cite{ beltagy2019scibert}, and BioMegatron \cite{shin2020biomegatron}, involved training on large publicly available medical corpora like PubMed or Semantic Scholar. Furthermore, LLMs of alternative architectures have also been employed, including BioELMo  \cite{jin2019probing}, BioBART \cite{BioBART}, and BioMed-RoBERTa \cite{gururangan2020don}. Another research direction capitalizes on the generalization capabilities of recent LLMs, where biomedical problems are addressed through prompting LLMs in zero-shot or few-shot settings. This approach has been utilized in various applications, such as medical report summarization  \cite{otmakhova2022m3}, medical writing \cite{biswas2023chatgpt}, and medical named entity recognition \cite{hu2023zero}, etc.

In this work, we propose \ours, a multi-level data with curated annotations at various granularity to comprehensively evaluate the strengths and limitations of LMs in healthcare.

\section{Dataset Design}
 
\ours~(shown in \cref{fig:dataset}) is designed with multiple NLU and NLG tasks at both the sentence and document levels, based on medical data collected from two different healthcare databases. Our data covers diverse combinations of human body parts and examination modalities. We first introduce the data sources where we collected the text input (\cref{sec: data composition}). Then  we present the expert-annotated ground truth labels\footnote{We use ``ground truth labels'' to represent both discriminative labels for NLU and golden sentences for NLG tasks.} created by our medical team\footnote{See description of the medical team in Appendix~\ref{appendix: medical_team}.} (\cref{sec: labels}).

\subsection{Input Data Composition} 
\label{sec: data composition}
\vspace{-0.em}
\paragraph{Sentence-Level Corpora} 
The sentence-level corpora used in this study are sourced from two well-constructed datasets: the sentence-level OpenI-annotated dataset~\citep{demner2016openi}, which consists of sentences from chest studies, and the VA-annotated dataset~\cite{he2023nothing}, which includes sentences about different body parts examined by different modalities.
These datasets have undergone de-identification, completion of missing terms and uniqueness checks. More details about the data preprocessing is given in Appendix~\ref{Appendix: data Preparation}. We use the officially released versions of the OpenI-annotated and VA-annotated datasets.
In addition, we provide new annotations for sentence-level tasks on these data sources.
\vspace{-0.5em}


\paragraph{Report-Level corpora} We collect the raw radiology reports from two distinct sources: (1) text corpus from MIMIC-CXR, which comprises records related to human chests \cite{johnson2019mimic}, (2) text corpus from the databases of a nationwide government healthcare system. 
We randomly collect data points about different body parts and exam modalities, resulting in multiple domains under different data distributions. The distribution of the domain is illustrated in \cref{fig: M4}. 
The collected data are processed with automatic de-identification, followed by a thorough human inspection to verify that no private information about patients or doctors is disclosed or hinted at in the text. We also employ an offline paraphrasing tool~\cite{prithivida2021parrot} to revise the text data collected from the second source. The paraphrasing is followed by another human inspection to filter out any unqualified records where the rewriting deviates significantly from the original report. The resulting data set can be considered ``synthesized'' and containing no privacy information but retaining realistic clinical conditions as the source data.

For each evaluation task, we split the data in a ratio of 7:1:2 for train/validate/test. 
 


\begin{table}[]
\centering
\resizebox{\linewidth}{!}{%
\begin{tabular}{@{}cccc@{}}
\toprule
\multicolumn{4}{c}{Disease Codes of \ours}                                     \\ \midrule
Enlarged Cardiomediastinum & Cardiomegaly  & Lung Opacity    & Lung Lesion  \\
Edema                      & Consolidation & Atelectasis     & Pneumothorax \\
Pleural Effusion           & Pleural Other & Support Devices & Pneumonia    \\
Dislocation                & Osteonecrosis & Fracture        & Gout         \\
Metatarsus Primus Varus    & Gas           & Swelling        & Psoriasis    \\
Enthesopathy               & Hammer Toe    & Osteomyelitis   & Mass         \\
Arthritis                  & Pes Planus    & Rheumatoid      & Cppd         \\
Hardware                   & Erosion       & Pes Cavus       & Coalition    \\
Subluxation                & Fracture      & Nodule          & Rupture      \\
Hallux Valgus              & Pneumonia     & Arthritis       & No Finding   \\ \bottomrule
\end{tabular}%
}
\caption{\small Report disease codes covered in \ours.}

\label{tab: medical codes}
\end{table}
\subsection{Expert Labels and Evaluation Tasks}
\label{sec: labels}

\subsection{Sentence-level Labels}
\paragraph{NLU Tasks} 
Identifying sentences with certain diagnostic properties is a practical use case in a real-world healthcare system. For example, identifying if a report sentence implies an abnormal finding about the patient or not. To test if language models can capture the medical semantics of single sentences, we first include abnormal sentence identification into our evaluation pool. We use the sentence-level corpora and the associated abnormality labels to classify abnormal sentences.  

Ambiguous sentences appear in radiology reports mainly due to the use of medical jargon whose meaning is different from daily usage, contradictory findings within the same sentence, or grammatical errors that mislead interpretation \cite{he2023nothing}. Accurate identification of such sentences is crucial, as they impede patients' comprehension of diagnostic decisions, leading to potential treatment delays and irreparable consequences. 
To the best of our knowledge, as a novel task proposed recently, current LMs may not readily include such a task into its pre-training stage. Therefore, evaluation of this task allows us to investigate how language models perform when the  tasks are unfamiliar. We leverage the report sentences and their associated ambiguous labels,
and our medical team re-examined and re-annotated the labels for ambiguous sentences.

\paragraph{NLG Task} 
Expanding beyond the previous ambiguous sentence identification, we include the task of sentence disambiguation as a sentence-level generation task. Proposed in \citet{he2023nothing}, sentence disambiguation aims to rewrite an ambiguous sentence in a way that its diagnostic findings are more explicitly expressed while at the same time, the original content of the report sentence is faithfully maintained. This requires rewritten sentences to avoid the change of the original pathological findings or introducing new findings. Similar to  ambiguous sentence identification, disambiguated rewriting presents a challenging generation task, not only because both the data and task formulation are not likely to be covered in the pre-training stage of existing language models, but also because there are two objectives that need to be optimized at the same time. In this task, based on the ambiguous sentences and their associated diagnostic labels,  our medical team manually created the disambiguated rewritings as the ground truth.

\subsection{Document-level Labels
}
\paragraph{NLU Task} 
To access if language models can capture the key findings of a radiology report, we consider Report Codes Prediction as an evaluation task.
This task involves categorizing reports into specific diagnostic codes based on the mentioned pathological findings. Therefore, different from sentence-level abnormality identification, this task requires a multi-label multi-class classification. Our medical team manually labels the medical codes of each report. Detailed information regarding the codes is provided in \cref{tab: medical codes}. More details about the expert-labeling procedure are provided in Appendix \ref{Appendix: data Preparation}.


\paragraph{NLG Task} 
Automatic medical summarization plays a crucial role in healthcare literature, by providing concise summaries, it saves time and manual effort for medical professionals when assessing the effectiveness of medical interventions. 
In our evaluation, we include report summarization as a task to assess the generation capability of language models.  The \textit{impression} section in each report serves as a summary that captures the supportive evidence for  clinical decisions. To ensure data quality, we conduct a manual inspection of all collected <report, impression> pairs, filtering out any pairs where the impression does not align with the corresponding report. It is worth noting that the curated parallel data of reports and summaries provide valuable support for future work in related fields.
\begin{table*}[h]
\centering
\resizebox{0.85\textwidth}{!}{%
\begin{tabular}{@{}cccccc@{}}
\toprule
\multicolumn{2}{c}{\multirow{2}{*}{Models}} &
  \multicolumn{2}{c}{Chest} &
  \multicolumn{2}{c}{Miscellaneous Domains} \\ \cmidrule(lr){3-4} \cmidrule(lr){5-6}
\multicolumn{2}{c}{}   & Abnormality $\uparrow$    & Ambiguity $\uparrow$        & Abnormality$\uparrow$       & Ambiguity $\uparrow$        \\ \midrule
\multirow{7}{*}{\begin{tabular}[c]{@{}c@{}}Adapted PLMs with \\ Fine-Tuning\end{tabular}} 
 & BERT             & 0.9791          & \textbf{0.9893} & 0.9607          & 0.9749          \\
 & RadBERT         & 0.9794          & 0.9869          & 0.9640          & \textbf{0.9813} \\
 & BioBERT         & 0.9791          & 0.9862          & 0.9614          & 0.9743          \\
 & ClinicalBERT     & \textbf{0.9809} & 0.9874          & 0.9588          & 0.9736          \\
 & BlueBERT         & 0.9803          & 0.9867          & 0.9601          & 0.9775          \\
 & BioMed-ReBERTa   & 0.9569          & 0.9758          & \textbf{0.9776} & 0.9788          \\ \cmidrule(lr){1-6}
\multirow{8}{*}{\begin{tabular}[c]{@{}c@{}}
LLMs Prompted by \\ Zero/Few Shot\end{tabular}} &
  zero-shot ChatGPT &
  0.9277 &
  0.6584 &
  0.8880 &
  0.5206 \\
 & few-shot ChatGPT    & 0.9498          & 0.5831          & 0.9099          & 0.5354          \\
 & zero-shot GPT-3     & 0.8762          & 0.8742          & 0.8243          & 0.6448          \\
 & few-shot GPT-3      & 0.9215          & 0.8320          & 0.9054          & 0.6371          \\  
 & zero-shot Vicuna-7B & 0.6987          & 0.2130          & 0.7261          & 0.3739          \\
 & few-shot Vicuna-7B  & 0.8071          & \underline{0.0785}          & 0.8166          & \underline{0.2844}          \\  \cmidrule(lr){2-6}
  & zero-shot BioMed LM &  \underline{0.6679 }& 0.3485 &  \underline{0.6273}         & 0.3726   \\
 & few-shot BioMedLM   &  0.7905  & 0.6804 & 0.7638           & 0.6804     \\\bottomrule
\end{tabular}%
}
\caption{\small Evaluation (accuracy) over two categories of PLMs on abnormality  identification and ambiguity identification tasks (sentence-level NLU). 
\textbf{Bold}: the  highest performance. \underline{Underlined}:  the lowest.
}
\label{tab:NLU-sentence}

\vspace{15pt}

\centering
\resizebox{\textwidth}{!}{%
\begin{tabular}{@{}cccccccc@{}}
\toprule
\multicolumn{2}{c}{\multirow{2}{*}{Model}} &
  \multicolumn{3}{c}{Chest} &
  \multicolumn{3}{c}{Miscellaneous  Domains} \\ \cmidrule(l){3-5}\cmidrule(l){6-8}
\multicolumn{2}{c}{} &
  \begin{tabular}[c]{@{}c@{}}Disambiguation \\ $\Delta Acc_{am} \uparrow$\end{tabular} &
  \begin{tabular}[c]{@{}c@{}}Content Distortion\\ $\Delta Acc_{ab} \downarrow$\end{tabular} &
  BLEU4 $\uparrow$  &
  \begin{tabular}[c]{@{}c@{}}Disambiguation \\ $\Delta Acc_{am} \uparrow$\end{tabular} &
  \begin{tabular}[c]{@{}c@{}}Content Distortion\\ $\Delta Acc_{ab} \downarrow$\end{tabular} &
  BLEU4 $\uparrow$  \\ \midrule
\multirow{4}{*}{\begin{tabular}[c]{@{}c@{}}Adapted PLMs \\ with \\ Fine-Tuning\end{tabular}} &
  Style Transfer &
  0.5010 &
  0.0510 &
  27.92 &
  0.3110 &
  0.2350 &
  31.17 \\
 &
  PPLM &
  0.3860 &
  0.1150 &
  57.88 &
  0.2700 &
  0.1460 &
  60.14 \\
 &
  DEPEN &
  0.5000 &
  0.0520 &
  60.48 &
  0.3530 &
  0.0470 &
  67.86 \\
 &
  MedDEPEN &
  0.4960 &
  0.0320 &
  57.88 &
  0.4810 &
  \textbf{0.0090} &
  68.88 \\ \midrule
\multirow{8}{*}{\begin{tabular}[c]{@{}c@{}}LLMs Prompted\\ by \\ Zero/Few Shots\end{tabular}} &
  zero-shot ChatGPT &
  0.6337 &
  \textbf{-0.0297} &
  60.73 &
  0.6539 &
  0.1483 &
  60.64 \\
 &
  few-shot ChatGPT &
  0.5875 &
  0.0000 &
  68.92 &
  0.6370 &
  0.0815 &
  67.98 \\
 &
  zero-shot GPT-3 &
  \textbf{0.6799} &
  -0.0132 &
  61.78 &
  \textbf{0.8022} &
  0.1528 &
  61.05 \\
 &
  few-shot GPT-3 &
  0.6139 &
  0.0000 &
  \textbf{76.33} &
  0.7146 &
  0.0607 &
  \textbf{77.09} \\
 &
  zero-shot Vicuna-7B &
  0.6230 &
  0.0693 &
  66.65 &
  0.6771 &
  \underline{0.3653} &
  64.64 \\
 &
  few-shot Vicuna-7B &
  0.5311 &
  \underline{0.1914} &
  62.55 &
  0.4811 &
  0.2739 &
  63.72 \\ \cmidrule(l){2-8} 
 &
  zero-shot BioMed LM &
  0.2211 &
  0.0066 &
  23.40 &
  \underline{0.1528} &
  0.1416 &
  24.11 \\
 &
  few-shot BioMed LM &
  \underline{0.1386} &
  -0.0262 &
  \underline{23.30} &
  0.3933 &
  0.3640 &
  \underline{23.48} \\ \bottomrule
\end{tabular}%
}
\caption{\small Evaluation on  disambiguated rewriting Tasks (sentence-level NLG). We report the disambiguation score, content distortion score (where smaller content distortion indicates higher fidelity), and BLEU4 score. 
\textbf{Bold}: the  best performance. \underline{Underlined}:  the worst.
}

\label{tab:NLG-sentence}
\end{table*}

\subsection{Evaluated Language Models}
We evaluate two categories of language models with \ours\footnote{The results presented are based on models evaluated as of the time of paper acceptance. We've added more results (e.g., on GPT4, LLaMa2, etc.) and will continue to include the newest ones in the Appendix \ref{appendix: more_results}.}:  (1)  domain-adapted pre-trained language models (Adapted PLMs), which are trainable models adapted on certain domain data, and (2)  general-purpose large language models (Prompted LLMs) which are used by zero/few-shot prompting. 

\subsubsection{Domain-adapted PLMs} Recent literature found it is effective to adapt pre-trained language models to certain narrow domains such as biomedical text by a continued training step on domain-specific data \cite{gururangan2020don}, following which we take a pre-trained (or generally adapted) language model, and test it on the \ours~test set. We also fine-tuned the models from this category to customize it to fit the tasks of \ours,  with their corresponding training data. For NLU tasks at both levels, we follow the evaluation setting of \citet{yan2022radbert} and investigate how: \textbf{BERTbase} \cite{devlin2018bert}, \textbf{RadBERT} \cite{yan2022radbert}, \textbf{BioBERT} \cite{lee2020biobert}, \textbf{clinicalBERT} \cite{huang2019clinicalbert}, \textbf{BlueBERT} \cite{peng2019transfer}, and \textbf{BioMed-ReBERTa-base} \cite{domains} perform on \ours. More details about those models are included in \cref{sec:related work}.


For the sentence-level NLG task, we follow the 
  the setting of \citet{he2023nothing} by evaluating: (1) \textbf{style transformer} \cite{dai-etal-2019-style} which transfers the original sentence into a less ambiguous style, (2) \textbf{PPLM} \cite{Dathathri2020Plug} which adds perturbation to LM to move the (re-)generation towards a less ambiguous direction, (3) \textbf{DEPEN} \cite{he-etal-2021-detect-perturb}  which is built upon PPLM and only re-generates ambiguous tokens detected before, and (4) \textbf{MedDEPEN} \cite{he2023nothing}, a biomedical-adapted DEPEN by introducing contrastive pre-training. Each work has included a transformer-based language model.  We refer the reader to the
original papers for more details.  

For the document-level NLG task, we follow the  setting of \citet{yan2022radbert} and customize previously adapted BERT-based models used before for the summarization task. 

\subsubsection{Prompted LLMs} 
We include the following general-purpose large language models to test their generalization in the healthcare domain: (1) \textbf{GPT3}: GPT-style large language models with 175B parameters \cite{brown2020language}. We use \texttt{davinci-003}.
(2) \textbf{ChatGPT}\footnote{\url{https://openai.com/blog/chatgpt}}: GPT-style large language model trained with Reinforcement Learning from Human Feedback (RLHF). We use \texttt{GPT3.5-turbo}. 
(3) \textbf{Vicuna-7B}~\cite{vicuna}: The finetuned version of LLaMa-7B~\cite{llama} with 70K user-shared ChatGPT conversations, which is capable of generating more detailed and well-structured answers.
(4) \textbf{BioMedLM}\footnote{\small \url{https://crfm.stanford.edu/2022/12/15/biomedlm.html}}: a 2.7B GPT-style language model trained exclusively on biomedical abstracts and papers from The Pile\cite{gao2020pile}.  

We prompt those LLMs under zero/few-shot settings, where we randomly select the examples from the training set of each task to compose prompts 5 times. We report the test results with the prompts which obtain optimal results on the validation set.
See Appendix \ref{Appendix: prompt} for more details.

\subsection{Evaluation Metrics}
For NLU tasks, we report classification metrics including accuracy and F1 scores. For NLG tasks, we report BLEU and ROUGE scores with respect to the ground truths labeled by our medical team. For sentence-level generation tasks (i.e., rewriting), to evaluate the objective of disambiguation, we follow the setting of \citet{he2023nothing} to report accuracy decrements of the ambiguity classifier ($\Delta Acc_{am}$) as the disambiguation metric. To evaluate the rewriting fidelity, we report the content distortion score, which is defined as the decrement of the accuracy from an abnormality classifier ($\Delta Acc_{ab}$). Therefore, higher distortion indicates a lower content fidelity.

\section{Results and Discussion}
\label{sec:Discussion}
\begin{table*}[h]
\centering
\resizebox{\textwidth}{!}{%
\begin{tabular}{@{}cccccccc@{}}
\toprule
\multicolumn{2}{c}{\multirow{2}{*}{Model}} &
  \multicolumn{2}{c}{Chest} &
  \multicolumn{2}{c}{Foot} &
  \multicolumn{2}{c}{Ankle} \\ \cmidrule(lr){3-4} \cmidrule(lr){5-6} \cmidrule(lr){7-8} 
\multicolumn{2}{c}{}   & avg Accuracy $\uparrow$     & avg EMR  $\uparrow$         & avg Accuracy & avg EMR  $\uparrow$    & avg Accuracy $\uparrow$ & avg EMR $\uparrow$  \\ \midrule
\multirow{6}{*}{\begin{tabular}[c]{@{}c@{}}Adapted PLMs \\ with \\ Fine-Tuning\end{tabular}} &
  BERT &
  0.8779 &
  0.2263 &
  \textbf{0.9754} &
  0.5635 & 
  0.9787 &
  0.6141\\
 & RadBERT         & 0.8785          & 0.1941          & 0.9710      &  0.4910      & 0.9773         & 0.5710       \\
 & BioBERT         & 0.8782          & \textbf{0.2400}          & 
  0.9750 &
  0.5617 &
  \textbf{0.9801} &
  \textbf{0.6266}       \\
 & ClinicalBERT     & 0.8780          & 0.2341          & 0.9731           & 0.5372         & 0.9798            & 0.6224       \\
 & BlueBERT        & \textbf{0.8843}          & 0.2380          & 0.9703          & \textbf{0.5939} & 
    0.9761 & 0.5752       \\
 & BioMed-ReBERTa   & 0.8579          & 0.1415          & 0.9692            & 0.4631          & 0.9752          & 0.5522    \\ 
 \cmidrule(lr){1-8} 
\multirow{8}{*}{\begin{tabular}[c]{@{}c@{}}LLMs Prompted\\ by \\ Zero/Few Shots\end{tabular}} &
  zero-shot ChatGPT &
  {0.5272} &
  0.1024 &
  0.9621 &
  0.4449 &
  0.9660 &
  0.4491 \\
 & few-shot ChatGPT    & 0.6485          & 0.1951          & 0.9621       & 0.4186     & 0.9690       & 0.4875  \\
 & zero-shot GPT-3     & \underline{0.2744}          & 0.1424          & 0.9621       & 0.4449     & \underline{0.1887}       & 0.6273  \\
 & few-shot GPT-3      & 0.8160          & 0.1805          & 0.9617       & 0.4186     & 0.9691       & 0.4908  \\
 & zero-shot Vicuna-7B & 0.8216            & \underline{0.0672}        & 0.9617       & 0.4186     & 0.9691       & 0.4908  \\  
 & few-shot Vicuna-7B  & 0.8228 & 0.0782 & \underline{0.5156 }      & \underline{0.1041}     & 0.9122       & \underline{0.4153}  \\ \cmidrule(l){2-8}
 & few-shot BioMed LM  & 0.8320   & 0.0689  & 0.9667   & 0.4664  & 0.9719 & 0.4980     \\ \bottomrule
\end{tabular}%
}
\caption{\small Evaluation on report codes prediction Task (Document-level NLU). We report the average accuracy over all classes of diseases and the exact match rate (EMR) between predictions and labels. \textbf{Bold}: the  highest performance. \underline{Underlined}:  the lowest. 
}
\label{tab:NLU-document}
\end{table*}

\begin{table*}[h]
\centering
\resizebox{0.7\textwidth}{!}{%
\begin{tabular}{@{}ccccccc@{}}
\toprule
\multicolumn{2}{c}{\multirow{2}{*}{Model}}                                                                       & \multicolumn{5}{c}{Miscellaneous Domains} \\ \cmidrule(l){3-7} 
\multicolumn{2}{c}{}   & ROUGE-1 $\uparrow$     & ROUGE-2 $\uparrow$     & ROUGE-L$\uparrow$   & Sum $\uparrow$             & BLEU4$\uparrow$    \\ \midrule
\multirow{8}{*}{\begin{tabular}[c]{@{}c@{}}Adapted PLMs \\ with \\ Fine-Tuning\end{tabular}} & BERT              & 20.48  & 7.46  & 18.57  & 46.52  & 30.28  \\
 & RadBERT            & 20.96      & 7.63       & 18.90   & 47.50          & 30.77 \\
 & BioBERT             & 20.79      & 7.62       & 18.78   & 47.19          & 30.49 \\
 & ClinicalBERT        & 21.22      & 7.85       & 19.18   & 48.26          & 30.81 \\
 & BlueBERT            & 20.83      & 7.78       & 18.90   & 47.51          & 30.93 \\
 & BioMed-ReBERTa      & 21.19      & 7.85       & 19.14   & 48.18          & 30.88 \\\midrule
\multirow{8}{*}{\begin{tabular}[c]{@{}c@{}}LLMs Prompted\\ by \\ Zero/Few Shots\end{tabular}} & zero-shot ChatGPT & 24.64  & 7.97  & 22.05  & 54.66  & 32.30  \\
 & few-shot ChatGPT    & \textbf{24.96 }     & 8.43       & \textbf{22.23}   & 55.62          & \textbf{35.43} \\
 & zero-shot GPT-3     & 24.06      & 8.67       & 21.52   & 54.24          & 24.43 \\
 & few-shot GPT-3      & 24.73      & \textbf{9.14}       & 22.16   & \textbf{56.03 }         & 34.72 \\
 & zero-shot Vicuna-7B & 20.93      & 6.96       & 18.71   & 46.60          & 20.94     \\
 & few-shot Vicuna-7B  & 21.42      & 7.26       & 19.22   & 47.90          & 22.00     \\ \cmidrule(l){2-7} 
 & zero-shot BioMed LM & 17.70      & 5.11       & 16.49   & 39.30          & \underline{15.43}    \\
 & few-shot BioMed LM  & \underline{12.15}      & \underline{3.50}       & \underline{11.22}   & \underline{26.87}          & 20.27    \\ \bottomrule
\end{tabular}%
}
\caption{\small Evaluation on report summarization task (Document-level NLG). We report the Rouge scores and BLEU4 scores. \textbf{Bold}: the  highest performance. \underline{Underlined}:  the lowest.}
\label{tab:NLG-document}
\end{table*}

\begin{figure}[h]
    \centering
    \begin{subfigure}[t]{0.45\linewidth}
        \centering
        \includegraphics[width=\textwidth]{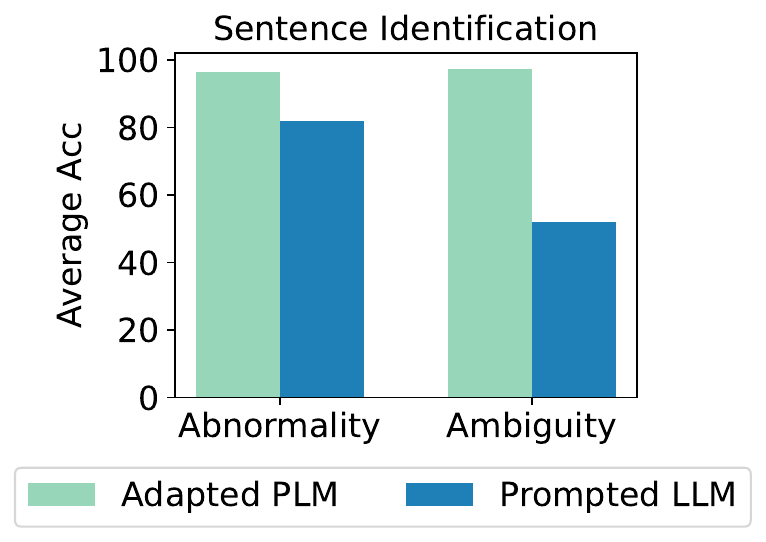}
        \caption{\centering \small Sentence-level NLU }
    \end{subfigure}%
    ~ 
     \centering
    \begin{subfigure}[t]{0.45\linewidth}
        \centering
        \includegraphics[width=\textwidth]{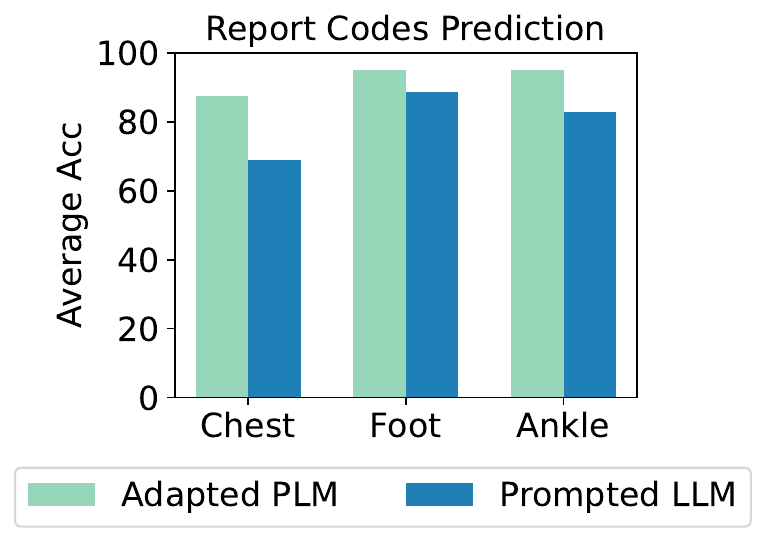}
        \caption{\centering \small Document-level NLU}
    \end{subfigure}%
    \hfill
    \begin{subfigure}[t]{0.45\linewidth}
        \centering
        \includegraphics[width=\textwidth]{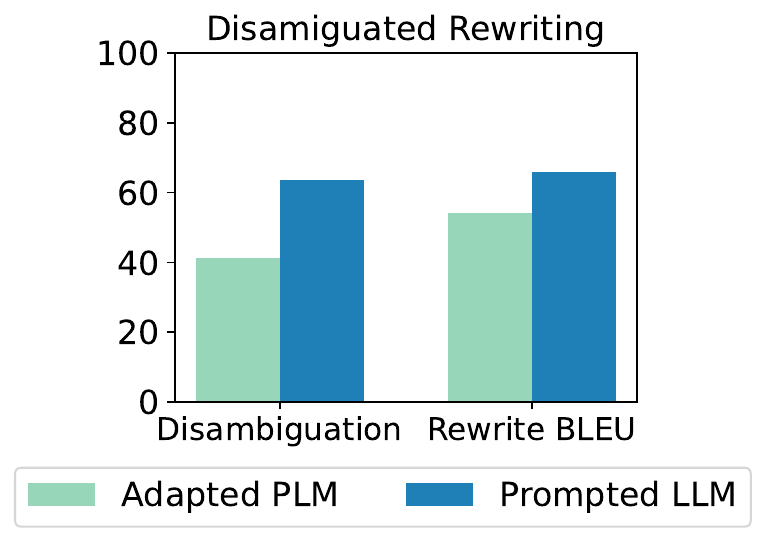}
        \caption{\small \centering Sentence-level NLG}
    \end{subfigure}
    \centering
    \begin{subfigure}[t]{0.45\linewidth}
        \centering
        \includegraphics[width=\textwidth]{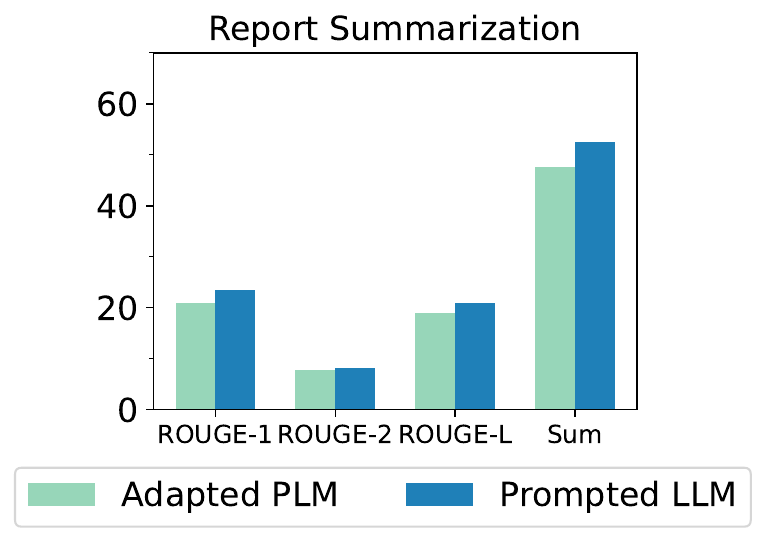}
        \caption{\small \centering  Document-level NLG}
    \end{subfigure}
    \caption{\small Average performance of adapted PLM and prompted LLM on different tasks and at different levels.}
    \label{fig: NLU vs NLG}
\end{figure}

In this section, We first present the results for sentence-level NLU tasks (Ambiguity Identification and Abnormality Identification) in \cref{tab:NLU-sentence}, then sentence-level NLG task (Disambiguated Rewriting) in \cref{tab:NLG-sentence}, finally document-level NLU (Code Prediction) and NLG (Report Summarization) tasks in \cref{tab:NLU-document} and \cref{tab:NLG-document}. 

\paragraph{The Effectiveness of Instruction Tuning }

While BioMed LM is the first large language model customized for the biomedical domain, we observe that it does not outperform adapted PLMs and most prompted LLMs in the majority of tasks. Particularly, BioMed LM has been found to be the weakest performer in tasks such as sentence identification, disambiguated rewriting, and report summarization.  We would like to highlight that, unlike other prompted LLMs such as ChatGPT, GPT-3, and Vicuna, BioMed LM lacks an Instruction Tuning step in its model training. This omission significantly impacts BioMed LM's ability to generate replies following the instructions from the given options. In zero-shot NLU tasks, only 40\% of the test cases receive appropriate responses at the sentence level and  the qualified rate drops to less than 1\% at the document level (so we did not report the results in \cref{tab:NLU-document}). In  few-shot report codes prediction, the document-based prompts often exceed BioMed LM's maximum threshold of 1024 tokens, resulting in query errors. In generation tasks, BioMed LM keeps returning irrelevant text. Our manual inspection reveals that the outputs rarely adhere to the given instructions in prompts or address the queries. This is further supported by the remarkably low BLEU or ROUGE scores in \cref{tab:NLG-sentence} and \cref{tab:NLG-document}. We provide more discussions in Appendix \ref{appendx:biomed}. These findings underscore the significance of Instruction Tuning and establish it as a crucial step when adapting prompted LLMs for specialized applications like healthcare decision-making.  

In the remainder of this section, we focus on addressing more intriguing questions based on average performance across a range of baselines (e.g., the average accuracy of adapted PLMs versus prompted LLMs), where we exclude BioMed LM from further consideration.

\paragraph{Discussion on Task Type and Granularity }

\begin{table}[ht]
\centering
\resizebox{\linewidth}{!}{%
\begin{tabular}{@{}ccccc@{}}
\toprule
\multirow{2}{*}{Model} & \multicolumn{2}{c}{Abnormality $\uparrow$} & \multicolumn{2}{c}{Ambiguity   $\uparrow$} \\ \cmidrule(lr){2-3} \cmidrule(lr){4-5} 
             & Chest  & Miscellaneous & Chest  & Miscellaneous \\ \midrule
Adapted PLM  & 0.9758 & 0.9526        & 0.9836 & 0.9621        \\
Prompted LLM & 0.8635 & 0.8451        & 0.5399 & 0.4893        \\ \bottomrule
\end{tabular}%
}
\caption{\small Average accuracy of adapted PLMs and prompted LLMs in NLU over different domains}
\label{tab:rare vs common domain}
\end{table}

In this section, we aim to determine the proficiency of language models at different levels and tasks. To achieve this, we begin by calculating the average accuracy scores of all adapted PLM baselines and prompted LLM baselines in sentence identification tasks. Similarly, we compute the average accuracy of adapted PLM and prompted LLM baselines in a document-level code classification task.

First, examining the results presented in \cref{fig: NLU vs NLG}, we observe that both adapted PLMs and prompted LLMs perform relatively similarly across different data levels.  However, it becomes apparent that adapted PLMs outperform prompted LLMs in NLU tasks, no matter whether it's on the sentence or document level. This suggests that fine-tuning provides a more effective means of injecting specific knowledge about narrow domains or tasks. On the other hand, consistently superior performance of prompted LLMs compared to adapted PLMs is observed in generation tasks, at both the sentence and document levels.  This can be attributed to multiple advantages of large-scale pre-training such as a larger model size or the benefits HFRL
in the LLMs we utilized, such as ChatGPT. These models demonstrate a capability to generate language that is more akin to human-like expressions, thereby achieving better generation scores.
These imply that fine-tuning PLM models can be a viable choice for NLU tasks, while prompting-based LLMs may be more suitable when healthcare professionals require an AI writer to help their work.


\begin{table}[ht]
\centering
\resizebox{\linewidth}{!}{%
\begin{tabular}{@{}cc|cc|cc@{}}
\toprule
\multirow{2}{*}{Model Family}     & \multirow{2}{*}{\# shot} & \multicolumn{2}{c|}{NLU (Accuracy$\uparrow$  )} & \multicolumn{2}{c}{NLG (BLEU$\uparrow$  )}  \\ \cmidrule{3-4}\cmidrule{5-6} 
                         &          & Individual & Average                 & Individual & Average                \\ \midrule
\multirow{2}{*}{ChatGPT} & 0-shot   & 0.78       & \multirow{2}{*}{0.79} & 51.22      & \multirow{2}{*}{47.15} \\
                         & Few-shot & 0.79       &                       & 43.08      &                        \\ \midrule
\multirow{2}{*}{GPT-3}   & 0-shot   & 0.66       & \multirow{2}{*}{0.76} & 49.08      & \multirow{2}{*}{55.90} \\
                         & Few-shot & 0.86       &                       & 62.71      &                        \\ \midrule
\multirow{2}{*}{Vicuna-7B} & 0-shot                   & 0.71 & \multirow{2}{*}{0.73} & 50.74 & \multirow{2}{*}{50.08} \\
                         & Few-shot & 0.75       &                       & 49.42      &                        \\ \midrule
\multirow{2}{*}{Average} & 0-shot   & \multicolumn{2}{c|}{0.72}           & \multicolumn{2}{c}{50.35}           \\
                         & Few-shot & \multicolumn{2}{c|}{0.80}           & \multicolumn{2}{c}{51.74}           \\ \bottomrule
\end{tabular}%
}
\caption{Average accuracy and BLEU of various LM families  with zero/few shots.}
\label{tab:LLM and fewshot}
\vspace{-8pt}

\end{table}

\paragraph{Common v.s. Rare Domains} In \cref{tab:rare vs common domain}, we explore the impact of the domain on language models in the healthcare field. We compute the average accuracy of adapted PLMs and prompted LLMs in abnormality identification  v.s.~ambiguity identification. We consistently observe higher performance from both adapted PLMs and prompted LLMs when working with data from the chest domain compared to miscellaneous domains. This superior performance can be attributed to the similarity between the chest data we tested and the pre-training data of the language models -- chest-related healthcare text is widely available in the public domain and can be included in the training corpus of PLMs. Similarly, LMs are expected to excel in abnormality identification tasks, which are a common research topic in current literature.

The most challenging scenario arises when both the data and task are unseen, specifically in the case of ambiguous identification within the miscellaneous domain. In such situations, there are limited or no examples available in the public domain.
Therefore, querying language models with (zero) few-shot learning proves to be less effective. 

\paragraph{Family of LLMs and Few Shot Learning} In this analysis, we examine the behavior of different language models (LLMs) with varying numbers of shots across different tasks. We calculate the average accuracy of ChatGPT, GPT3, and Vicuna-7B in NLU tasks and the average BLEU scores in NLG tasks. Additionally, we consider the average performance achieved in zero-shot or few-shot settings (\cref{tab:LLM and fewshot}). From the table, it is evident that in most cases, providing additional examples assists LMs in making predictions for NLU tasks. However, in NLG tasks, no consistent trend is observed, indicating the need for further research to discover optimal prompts. We do not observe a clear advantage of any specific LLM family over others, suggesting that the choice of the optimal LLM family for a given task may vary on a case-by-case basis.

\section{Conclusion} We introduce \ours, a multi-task, multi-level, and multi-domain medical benchmark designed to serve as a comprehensive testbed for advanced language models.  Through extensive evaluation experiments, we thoroughly analyze the capabilities and limitations of current LLMs in tackling various medical tasks, such as the effectiveness of instruction tuning and the performance disparities between adapted and prompted LMs in NLU and NLG tasks. 
Our findings provide valuable insights and serve as a handbook for future research in utilizing LLMs to enhance healthcare practices. 
    
\section{Limitations}
In our efforts to provide a comprehensive testbed for current advanced language models, we have included multiple tasks. However, we acknowledge that there may be other tasks of interest that could have been analyzed, such as medical named entity recognition, multi-document report summarization, etc. We plan to expand the range of test tasks in future iterations of the \ours~benchmark. We'd like to note that due to computing constraints, we were unable to evaluate some large language models such as Vicuna-60B or OPT-175B \cite{zhang2022opt}. Our evaluation was focused on  popular large language models with reasonably large sizes. In future work, we consider addressing this limitation by incorporating these larger language models into our testbed.

\section{Ethics Statement}
Our data underwent a rigorous de-identification process and were carefully reviewed by human evaluators following strict anonymization criteria. Moreover, the collection of data from real-world healthcare systems has received the necessary IRB approval (VASDHS IRB protocol 200086), ensuring compliance with ethical standards. To further ensure ethical usage, before inputting the data into large language models, including commercial ones like ChatGPT, we conducted data synthesis and subjected it to additional human inspection. These steps were taken to address any potential ethical concerns associated with the data.
    
It is important to highlight that the responsible and safe usage of biomedical data is a critical requirement in AI for healthcare, especially in the use case of large language models which are noticed to suffer from different kinds of potential harms \cite{leino2018featurewise,  lloyd2018bias, he2021detect,xu2022leashing, he-etal-2022-controlling} and weaknesses \cite{ribeiro2020beyond, stuart2018microsoft,  he2023targeted}. Therefore, we strongly recommend that our benchmark be used in conjunction with expert auditing to ensure the highest level of safety in real-world applications.

\subsection*{Acknowledgments}
The authors would like to thank NVIDIA corporation for the award of two RTX A6000 GPUs to CNH through the Applied Research Accelerator Program.

\bibliography{emnlp2023}

\begin{thebibliography}{55}
\expandafter\ifx\csname natexlab\endcsname\relax\def\natexlab#1{#1}\fi

\bibitem[{Ammar et~al.(2018)Ammar, Groeneveld, Bhagavatula, Beltagy, Crawford,
  Downey, Dunkelberger, Elgohary, Feldman, Ha et~al.}]{ammar2018construction}
Waleed Ammar, Dirk Groeneveld, Chandra Bhagavatula, Iz~Beltagy, Miles Crawford,
  Doug Downey, Jason Dunkelberger, Ahmed Elgohary, Sergey Feldman, Vu~Ha,
  et~al. 2018.
\newblock Construction of the literature graph in semantic scholar.
\newblock In \emph{Proceedings of the 2018 Conference of the North American
  Chapter of the Association for Computational Linguistics: Human Language
  Technologies, Volume 3 (Industry Papers)}, pages 84--91.

\bibitem[{Banarescu et~al.(2013)Banarescu, Bonial, Cai, Georgescu, Griffitt,
  Hermjakob, Knight, Koehn, Palmer, and Schneider}]{banarescu2013abstract}
Laura Banarescu, Claire Bonial, Shu Cai, Madalina Georgescu, Kira Griffitt, Ulf
  Hermjakob, Kevin Knight, Philipp Koehn, Martha Palmer, and Nathan Schneider.
  2013.
\newblock Abstract meaning representation for sembanking.
\newblock In \emph{Proceedings of the 7th linguistic annotation workshop and
  interoperability with discourse}, pages 178--186.

\bibitem[{Beltagy et~al.(2019)Beltagy, Lo, and Cohan}]{beltagy2019scibert}
Iz~Beltagy, Kyle Lo, and Arman Cohan. 2019.
\newblock Scibert: A pretrained language model for scientific text.
\newblock \emph{arXiv preprint arXiv:1903.10676}.

\bibitem[{Biswas(2023)}]{biswas2023chatgpt}
Som Biswas. 2023.
\newblock Chatgpt and the future of medical writing.

\bibitem[{Brown et~al.(2020)Brown, Mann, Ryder, Subbiah, Kaplan, Dhariwal,
  Neelakantan, Shyam, Sastry, Askell et~al.}]{brown2020language}
Tom Brown, Benjamin Mann, Nick Ryder, Melanie Subbiah, Jared~D Kaplan, Prafulla
  Dhariwal, Arvind Neelakantan, Pranav Shyam, Girish Sastry, Amanda Askell,
  et~al. 2020.
\newblock Language models are few-shot learners.
\newblock \emph{Advances in neural information processing systems},
  33:1877--1901.

\bibitem[{Charlton et~al.(2022)Charlton, Kotzen, Mej{\'\i}a-Mej{\'\i}a, Aston,
  Budidha, Mant, Pettit, Behar, and Kyriacou}]{charlton2022detecting}
Peter~H Charlton, Kevin Kotzen, Elisa Mej{\'\i}a-Mej{\'\i}a, Philip~J Aston,
  Karthik Budidha, Jonathan Mant, Callum Pettit, Joachim~A Behar, and Panicos~A
  Kyriacou. 2022.
\newblock Detecting beats in the photoplethysmogram: benchmarking open-source
  algorithms.
\newblock \emph{Physiological Measurement}, 43(8):085007.

\bibitem[{Chiang et~al.(2023)Chiang, Li, Lin, Sheng, Wu, Zhang, Zheng, Zhuang,
  Zhuang, Gonzalez, Stoica, and Xing}]{vicuna}
Wei-Lin Chiang, Zhuohan Li, Zi~Lin, Ying Sheng, Zhanghao Wu, Hao Zhang, Lianmin
  Zheng, Siyuan Zhuang, Yonghao Zhuang, Joseph~E. Gonzalez, Ion Stoica, and
  Eric~P. Xing. 2023.
\newblock \href {https://vicuna.lmsys.org} {Vicuna: An open-source chatbot
  impressing gpt-4 with 90\%* chatgpt quality}.

\bibitem[{Dai et~al.(2019)Dai, Liang, Qiu, and Huang}]{dai-etal-2019-style}
Ning Dai, Jianze Liang, Xipeng Qiu, and Xuanjing Huang. 2019.
\newblock \href {https://doi.org/10.18653/v1/P19-1601} {Style transformer:
  Unpaired text style transfer without disentangled latent representation}.
\newblock In \emph{Proceedings of the 57th Annual Meeting of the Association
  for Computational Linguistics}, pages 5997--6007, Florence, Italy.
  Association for Computational Linguistics.

\bibitem[{Damodaran(2021)}]{prithivida2021parrot}
Prithiviraj Damodaran. 2021.
\newblock Parrot: Paraphrase generation for {NLU}.
\newblock \url{https://github.com/PrithivirajDamodaran/Parrot_Paraphraser}.

\bibitem[{Dathathri et~al.(2020)Dathathri, Madotto, Lan, Hung, Frank, Molino,
  Yosinski, and Liu}]{Dathathri2020Plug}
Sumanth Dathathri, Andrea Madotto, Janice Lan, Jane Hung, Eric Frank, Piero
  Molino, Jason Yosinski, and Rosanne Liu. 2020.
\newblock \href {https://openreview.net/forum?id=H1edEyBKDS} {Plug and play
  language models: A simple approach to controlled text generation}.
\newblock In \emph{International Conference on Learning Representations}.

\bibitem[{Demner-Fushman et~al.(2016)Demner-Fushman, Kohli, Rosenman, Shooshan,
  Rodriguez, Antani, Thoma, and McDonald}]{demner2016openi}
Dina Demner-Fushman, Marc~D Kohli, Marc~B Rosenman, Sonya~E Shooshan, Laritza
  Rodriguez, Sameer Antani, George~R Thoma, and Clement~J McDonald. 2016.
\newblock Preparing a collection of radiology examinations for distribution and
  retrieval.
\newblock \emph{Journal of the American Medical Informatics Association},
  23(2):304--310.

\bibitem[{Dernoncourt and Lee(2017)}]{dernoncourt2017pubmed}
Franck Dernoncourt and Ji~Young Lee. 2017.
\newblock Pubmed 200k rct: a dataset for sequential sentence classification in
  medical abstracts.
\newblock \emph{IJCNLP 2017}, page 308.

\bibitem[{Devlin et~al.(2018)Devlin, Chang, Lee, and
  Toutanova}]{devlin2018bert}
Jacob Devlin, Ming-Wei Chang, Kenton Lee, and Kristina Toutanova. 2018.
\newblock Bert: Pre-training of deep bidirectional transformers for language
  understanding.
\newblock \emph{arXiv preprint arXiv:1810.04805}.

\bibitem[{Edin et~al.(2023)Edin, Junge, Havtorn, Borgholt, Maistro, Ruotsalo,
  and Maal{\o}e}]{edin2023automated}
Joakim Edin, Alexander Junge, Jakob~D Havtorn, Lasse Borgholt, Maria Maistro,
  Tuukka Ruotsalo, and Lars Maal{\o}e. 2023.
\newblock Automated medical coding on mimic-iii and mimic-iv: A critical review
  and replicability study.
\newblock \emph{arXiv preprint arXiv:2304.10909}.

\bibitem[{Gao et~al.(2020)Gao, Biderman, Black, Golding, Hoppe, Foster, Phang,
  He, Thite, Nabeshima et~al.}]{gao2020pile}
Leo Gao, Stella Biderman, Sid Black, Laurence Golding, Travis Hoppe, Charles
  Foster, Jason Phang, Horace He, Anish Thite, Noa Nabeshima, et~al. 2020.
\newblock The pile: An 800gb dataset of diverse text for language modeling.
\newblock \emph{arXiv preprint arXiv:2101.00027}.

\bibitem[{Guo et~al.(2023)Guo, Wang, Wang, and Yu}]{guo2023dr}
Zhen Guo, Peiqi Wang, Yanwei Wang, and Shangdi Yu. 2023.
\newblock Dr. llama: Improving small language models in domain-specific qa via
  generative data augmentation.
\newblock \emph{arXiv preprint arXiv:2305.07804}.

\bibitem[{Gupta et~al.(2021)Gupta, Bharti, Nokhiz, and
  Karnick}]{gupta2021sumpubmed}
Vivek Gupta, Prerna Bharti, Pegah Nokhiz, and Harish Karnick. 2021.
\newblock Sumpubmed: Summarization dataset of pubmed scientific articles.
\newblock In \emph{Proceedings of the 59th Annual Meeting of the Association
  for Computational Linguistics and the 11th International Joint Conference on
  Natural Language Processing: Student Research Workshop}, pages 292--303.

\bibitem[{Gururangan et~al.(2020{\natexlab{a}})Gururangan, Marasovi{\'c},
  Swayamdipta, Lo, Beltagy, Downey, and Smith}]{gururangan2020don}
Suchin Gururangan, Ana Marasovi{\'c}, Swabha Swayamdipta, Kyle Lo, Iz~Beltagy,
  Doug Downey, and Noah~A Smith. 2020{\natexlab{a}}.
\newblock Don’t stop pretraining: Adapt language models to domains and tasks.
\newblock In \emph{Proceedings of the 58th Annual Meeting of the Association
  for Computational Linguistics}, pages 8342--8360.

\bibitem[{Gururangan et~al.(2020{\natexlab{b}})Gururangan, Marasović,
  Swayamdipta, Lo, Beltagy, Downey, and Smith}]{domains}
Suchin Gururangan, Ana Marasović, Swabha Swayamdipta, Kyle Lo, Iz~Beltagy,
  Doug Downey, and Noah~A. Smith. 2020{\natexlab{b}}.
\newblock Don't stop pretraining: Adapt language models to domains and tasks.
\newblock In \emph{Proceedings of ACL}.

\bibitem[{He et~al.(2021{\natexlab{a}})He, Majumder, and
  McAuley}]{he-etal-2021-detect-perturb}
Zexue He, Bodhisattwa~Prasad Majumder, and Julian McAuley. 2021{\natexlab{a}}.
\newblock \href {https://doi.org/10.18653/v1/2021.findings-emnlp.352} {Detect
  and perturb: Neutral rewriting of biased and sensitive text via
  gradient-based decoding}.
\newblock In \emph{Findings of the Association for Computational Linguistics:
  EMNLP 2021}, pages 4173--4181, Punta Cana, Dominican Republic. Association
  for Computational Linguistics.

\bibitem[{He et~al.(2021{\natexlab{b}})He, Majumder, and
  McAuley}]{he2021detect}
Zexue He, Bodhisattwa~Prasad Majumder, and Julian McAuley. 2021{\natexlab{b}}.
\newblock Detect and perturb: Neutral rewriting of biased and sensitive text
  via gradient-based decoding.
\newblock In \emph{Findings of the Association for Computational Linguistics:
  EMNLP 2021}, pages 4173--4181.

\bibitem[{He et~al.(2023{\natexlab{a}})He, Ribeiro, and Khani}]{he2023targeted}
Zexue He, Marco~Tulio Ribeiro, and Fereshte Khani. 2023{\natexlab{a}}.
\newblock Targeted data generation: Finding and fixing model weaknesses.
\newblock \emph{arXiv preprint arXiv:2305.17804}.

\bibitem[{He et~al.(2022)He, Wang, McAuley, and
  Majumder}]{he-etal-2022-controlling}
Zexue He, Yu~Wang, Julian McAuley, and Bodhisattwa~Prasad Majumder. 2022.
\newblock \href {https://doi.org/10.18653/v1/2022.findings-emnlp.431}
  {Controlling bias exposure for fair interpretable predictions}.
\newblock In \emph{Findings of the Association for Computational Linguistics:
  EMNLP 2022}, pages 5854--5866, Abu Dhabi, United Arab Emirates. Association
  for Computational Linguistics.

\bibitem[{He et~al.(2023{\natexlab{b}})He, Yan, Gentili, McAuley, and
  Hsu}]{he2023nothing}
Zexue He, An~Yan, Amilcare Gentili, Julian McAuley, and Chun-Nan Hsu.
  2023{\natexlab{b}}.
\newblock ``{N}othing abnormal'': Disambiguating medical reports via
  contrastive knowledge infusion.
\newblock In \emph{Proceedings of the 37th {AAAI} Conference on Artificial
  Intelligence}.
\newblock ArXiv preprint arXiv:2305.08300.

\bibitem[{Hu et~al.(2023)Hu, Ameer, Zuo, Peng, Zhou, Li, Li, Li, Jiang, and
  Xu}]{hu2023zero}
Yan Hu, Iqra Ameer, Xu~Zuo, Xueqing Peng, Yujia Zhou, Zehan Li, Yiming Li,
  Jianfu Li, Xiaoqian Jiang, and Hua Xu. 2023.
\newblock Zero-shot clinical entity recognition using chatgpt.
\newblock \emph{arXiv preprint arXiv:2303.16416}.

\bibitem[{Huang et~al.(2019)Huang, Altosaar, and
  Ranganath}]{huang2019clinicalbert}
Kexin Huang, Jaan Altosaar, and Rajesh Ranganath. 2019.
\newblock Clinicalbert: Modeling clinical notes and predicting hospital
  readmission.
\newblock \emph{arXiv preprint arXiv:1904.05342}.

\bibitem[{Irvin et~al.(2019)Irvin, Rajpurkar, Ko, Yu, Ciurea-Ilcus, Chute,
  Marklund, Haghgoo, Ball, Shpanskaya et~al.}]{irvin2019chexpert}
Jeremy Irvin, Pranav Rajpurkar, Michael Ko, Yifan Yu, Silviana Ciurea-Ilcus,
  Chris Chute, Henrik Marklund, Behzad Haghgoo, Robyn Ball, Katie Shpanskaya,
  et~al. 2019.
\newblock Chexpert: A large chest radiograph dataset with uncertainty labels
  and expert comparison.
\newblock In \emph{Proceedings of the AAAI conference on artificial
  intelligence}, volume~33, pages 590--597.

\bibitem[{Jin et~al.(2019{\natexlab{a}})Jin, Dhingra, Cohen, and
  Lu}]{jin2019probing}
Qiao Jin, Bhuwan Dhingra, William~W Cohen, and Xinghua Lu. 2019{\natexlab{a}}.
\newblock Probing biomedical embeddings from language models.
\newblock \emph{NAACL HLT 2019}, page~82.

\bibitem[{Jin et~al.(2019{\natexlab{b}})Jin, Dhingra, Liu, Cohen, and
  Lu}]{jin2019pubmedqa}
Qiao Jin, Bhuwan Dhingra, Zhengping Liu, William Cohen, and Xinghua Lu.
  2019{\natexlab{b}}.
\newblock Pubmedqa: A dataset for biomedical research question answering.
\newblock In \emph{Proceedings of the 2019 Conference on Empirical Methods in
  Natural Language Processing and the 9th International Joint Conference on
  Natural Language Processing (EMNLP-IJCNLP)}, pages 2567--2577.

\bibitem[{Johnson et~al.(2023)Johnson, Bulgarelli, Shen, Gayles, Shammout,
  Horng, Pollard, Moody, Gow, Lehman et~al.}]{johnson2023mimic}
Alistair~EW Johnson, Lucas Bulgarelli, Lu~Shen, Alvin Gayles, Ayad Shammout,
  Steven Horng, Tom~J Pollard, Benjamin Moody, Brian Gow, Li-wei~H Lehman,
  et~al. 2023.
\newblock Mimic-iv, a freely accessible electronic health record dataset.
\newblock \emph{Scientific data}, 10(1):1.

\bibitem[{Johnson et~al.(2019)Johnson, Pollard, Greenbaum, Lungren, Deng, Peng,
  Lu, Mark, Berkowitz, and Horng}]{johnson2019mimic}
Alistair~EW Johnson, Tom~J Pollard, Nathaniel~R Greenbaum, Matthew~P Lungren,
  Chih-ying Deng, Yifan Peng, Zhiyong Lu, Roger~G Mark, Seth~J Berkowitz, and
  Steven Horng. 2019.
\newblock Mimic-cxr-jpg, a large publicly available database of labeled chest
  radiographs.
\newblock \emph{arXiv preprint arXiv:1901.07042}.

\bibitem[{Johnson et~al.(2016)Johnson, Pollard, Shen, Lehman, Feng, Ghassemi,
  Moody, Szolovits, Anthony~Celi, and Mark}]{johnson2016mimic}
Alistair~EW Johnson, Tom~J Pollard, Lu~Shen, Li-wei~H Lehman, Mengling Feng,
  Mohammad Ghassemi, Benjamin Moody, Peter Szolovits, Leo Anthony~Celi, and
  Roger~G Mark. 2016.
\newblock Mimic-iii, a freely accessible critical care database.
\newblock \emph{Scientific data}, 3(1):1--9.

\bibitem[{Lee et~al.(2020)Lee, Yoon, Kim, Kim, Kim, So, and
  Kang}]{lee2020biobert}
Jinhyuk Lee, Wonjin Yoon, Sungdong Kim, Donghyeon Kim, Sunkyu Kim, Chan~Ho So,
  and Jaewoo Kang. 2020.
\newblock Biobert: a pre-trained biomedical language representation model for
  biomedical text mining.
\newblock \emph{Bioinformatics}, 36(4):1234--1240.

\bibitem[{Leino et~al.(2019)Leino, Fredrikson, Black, Sen, and
  Datta}]{leino2018featurewise}
Klas Leino, Matt Fredrikson, Emily Black, Shayak Sen, and Anupam Datta. 2019.
\newblock \href {https://openreview.net/forum?id=S1ecm2C9K7} {Feature-wise bias
  amplification}.
\newblock In \emph{International Conference on Learning Representations}.

\bibitem[{Lloyd(2018)}]{lloyd2018bias}
Kirsten Lloyd. 2018.
\newblock Bias amplification in artificial intelligence systems.
\newblock \emph{arXiv preprint arXiv:1809.07842}.

\bibitem[{Luo et~al.(2022)Luo, Sun, Xia, Qin, Zhang, Poon, and
  Liu}]{10.1093/bib/bbac409}
Renqian Luo, Liai Sun, Yingce Xia, Tao Qin, Sheng Zhang, Hoifung Poon, and
  Tie-Yan Liu. 2022.
\newblock \href {https://doi.org/10.1093/bib/bbac409} {{BioGPT: generative
  pre-trained transformer for biomedical text generation and mining}}.
\newblock \emph{Briefings in Bioinformatics}, 23(6).
\newblock Bbac409.

\bibitem[{McCreery et~al.(2020)McCreery, Katariya, Kannan, Chablani, and
  Amatriain}]{mccreery2020effective}
Clara~H McCreery, Namit Katariya, Anitha Kannan, Manish Chablani, and Xavier
  Amatriain. 2020.
\newblock Effective transfer learning for identifying similar questions:
  matching user questions to covid-19 faqs.
\newblock In \emph{Proceedings of the 26th ACM SIGKDD international conference
  on knowledge discovery \& data mining}, pages 3458--3465.

\bibitem[{Otmakhova et~al.(2022)Otmakhova, Verspoor, Baldwin, Yepes, and
  Lau}]{otmakhova2022m3}
Julia Otmakhova, Karin Verspoor, Timothy Baldwin, Antonio~Jimeno Yepes, and
  Jey~Han Lau. 2022.
\newblock M3: Multi-level dataset for multi-document summarisation of medical
  studies.
\newblock In \emph{Findings of the Association for Computational Linguistics:
  EMNLP 2022}, pages 3887--3901.

\bibitem[{Pampari et~al.(2018)Pampari, Raghavan, Liang, and
  Peng}]{pampari2018emrqa}
Anusri Pampari, Preethi Raghavan, Jennifer Liang, and Jian Peng. 2018.
\newblock emrqa: A large corpus for question answering on electronic medical
  records.
\newblock In \emph{Proceedings of the 2018 Conference on Empirical Methods in
  Natural Language Processing}, pages 2357--2368.

\bibitem[{Pappas et~al.(2018)Pappas, Androutsopoulos, and
  Papageorgiou}]{pappas2018bioread}
Dimitris Pappas, Ion Androutsopoulos, and Harris Papageorgiou. 2018.
\newblock Bioread: A new dataset for biomedical reading comprehension.
\newblock In \emph{Proceedings of the Eleventh International Conference on
  Language Resources and Evaluation (LREC 2018)}.

\bibitem[{Peng et~al.(2019)Peng, Yan, and Lu}]{peng2019transfer}
Yifan Peng, Shankai Yan, and Zhiyong Lu. 2019.
\newblock Transfer learning in biomedical natural language processing: An
  evaluation of bert and elmo on ten benchmarking datasets.
\newblock In \emph{Proceedings of the 2019 Workshop on Biomedical Natural
  Language Processing (BioNLP 2019)}, pages 58--65.

\bibitem[{Ribeiro et~al.(2020)Ribeiro, Wu, Guestrin, and
  Singh}]{ribeiro2020beyond}
Marco~Tulio Ribeiro, Tongshuang Wu, Carlos Guestrin, and Sameer Singh. 2020.
\newblock Beyond accuracy: Behavioral testing of nlp models with checklist.
\newblock In \emph{Proceedings of the 58th Annual Meeting of the Association
  for Computational Linguistics}, pages 4902--4912.

\bibitem[{Shin et~al.(2020)Shin, Zhang, Bakhturina, Puri, Patwary, Shoeybi, and
  Mani}]{shin2020biomegatron}
Hoo-Chang Shin, Yang Zhang, Evelina Bakhturina, Raul Puri, Mostofa Patwary,
  Mohammad Shoeybi, and Raghav Mani. 2020.
\newblock Biomegatron: Larger biomedical domain language model.
\newblock In \emph{Proceedings of the 2020 Conference on Empirical Methods in
  Natural Language Processing (EMNLP)}, pages 4700--4706.

\bibitem[{Stuart-Ulin(2018)}]{stuart2018microsoft}
Chloe~Rose Stuart-Ulin. 2018.
\newblock Microsoft’s politically correct chatbot is even worse than its
  racist one.
\newblock \emph{Quartz Ideas}, 31.

\bibitem[{Tang et~al.(2023)Tang, Han, Jiang, and Hu}]{tang2023does}
Ruixiang Tang, Xiaotian Han, Xiaoqian Jiang, and Xia Hu. 2023.
\newblock Does synthetic data generation of llms help clinical text mining?
\newblock \emph{arXiv preprint arXiv:2303.04360}.

\bibitem[{Touvron et~al.(2023{\natexlab{a}})Touvron, Lavril, Izacard, Martinet,
  Lachaux, Lacroix, Rozi{\`e}re, Goyal, Hambro, Azhar, Rodriguez, Joulin,
  Grave, and Lample}]{llama}
Hugo Touvron, Thibaut Lavril, Gautier Izacard, Xavier Martinet, Marie-Anne
  Lachaux, Timoth{\'e}e Lacroix, Baptiste Rozi{\`e}re, Naman Goyal, Eric
  Hambro, Faisal Azhar, Aurelien Rodriguez, Armand Joulin, Edouard Grave, and
  Guillaume Lample. 2023{\natexlab{a}}.
\newblock Llama: Open and efficient foundation language models.
\newblock \emph{arXiv preprint arXiv:2302.13971}.

\bibitem[{Touvron et~al.(2023{\natexlab{b}})Touvron, Martin, Stone, Albert,
  Almahairi, Babaei, Bashlykov, Batra, Bhargava, Bhosale
  et~al.}]{touvron2023llama}
Hugo Touvron, Louis Martin, Kevin Stone, Peter Albert, Amjad Almahairi, Yasmine
  Babaei, Nikolay Bashlykov, Soumya Batra, Prajjwal Bhargava, Shruti Bhosale,
  et~al. 2023{\natexlab{b}}.
\newblock Llama 2: Open foundation and fine-tuned chat models.
\newblock \emph{arXiv preprint arXiv:2307.09288}.

\bibitem[{Tsatsaronis et~al.(2015)Tsatsaronis, Balikas, Malakasiotis, Partalas,
  Zschunke, Alvers, Weissenborn, Krithara, Petridis, Polychronopoulos
  et~al.}]{tsatsaronis2015overview}
George Tsatsaronis, Georgios Balikas, Prodromos Malakasiotis, Ioannis Partalas,
  Matthias Zschunke, Michael~R Alvers, Dirk Weissenborn, Anastasia Krithara,
  Sergios Petridis, Dimitris Polychronopoulos, et~al. 2015.
\newblock An overview of the bioasq large-scale biomedical semantic indexing
  and question answering competition.
\newblock \emph{BMC bioinformatics}, 16(1):1--28.

\bibitem[{Wu et~al.(2023)Wu, Lin, Zhang, Zhang, Wang, and Xie}]{wu2023pmcllama}
Chaoyi Wu, Weixiong Lin, Xiaoman Zhang, Ya~Zhang, Yanfeng Wang, and Weidi Xie.
  2023.
\newblock \href {http://arxiv.org/abs/2304.14454} {Pmc-llama: Towards building
  open-source language models for medicine}.

\bibitem[{Xu et~al.(2022)Xu, He, He, and McAuley}]{xu2022leashing}
Canwen Xu, Zexue He, Zhankui He, and Julian McAuley. 2022.
\newblock Leashing the inner demons: Self-detoxification for language models.
\newblock In \emph{Proceedings of the AAAI Conference on Artificial
  Intelligence}, volume~36, pages 11530--11537.

\bibitem[{Yan et~al.(2021{\natexlab{a}})Yan, He, Lu, Du, Chang, Gentili,
  McAuley, and Hsu}]{yan2021weakly}
An~Yan, Zexue He, Xing Lu, Jiang Du, Eric Chang, Amilcare Gentili, Julian
  McAuley, and Chun-Nan Hsu. 2021{\natexlab{a}}.
\newblock Weakly supervised contrastive learning for chest x-ray report
  generation.
\newblock \emph{arXiv preprint arXiv:2109.12242}.

\bibitem[{Yan et~al.(2021{\natexlab{b}})Yan, He, Lu, Du, Chang, Gentili,
  McAuley, and Hsu}]{yan-etal-2021-weakly-supervised}
An~Yan, Zexue He, Xing Lu, Jiang Du, Eric Chang, Amilcare Gentili, Julian
  McAuley, and Chun-Nan Hsu. 2021{\natexlab{b}}.
\newblock \href {https://doi.org/10.18653/v1/2021.findings-emnlp.336} {Weakly
  supervised contrastive learning for chest {X}-ray report generation}.
\newblock In \emph{Findings of the Association for Computational Linguistics:
  EMNLP 2021}, pages 4009--4015, Punta Cana, Dominican Republic. Association
  for Computational Linguistics.

\bibitem[{Yan et~al.(2022)Yan, McAuley, Lu, Du, Chang, Gentili, and
  Hsu}]{yan2022radbert}
An~Yan, Julian McAuley, Xing Lu, Jiang Du, Eric~Y Chang, Amilcare Gentili, and
  Chun-Nan Hsu. 2022.
\newblock Radbert: Adapting transformer-based language models to radiology.
\newblock \emph{Radiology: Artificial Intelligence}, 4(4):e210258.

\bibitem[{Yuan et~al.(2022)Yuan, Yuan, Gan, Zhang, Xie, and Yu}]{BioBART}
Hongyi Yuan, Zheng Yuan, Ruyi Gan, Jiaxing Zhang, Yutao Xie, and Sheng Yu.
  2022.
\newblock \href {http://arxiv.org/abs/2204.03905} {Biobart: Pretraining and
  evaluation of a biomedical generative language model}.

\bibitem[{Zhang et~al.(2022)Zhang, Roller, Goyal, Artetxe, Chen, Chen, Dewan,
  Diab, Li, Lin et~al.}]{zhang2022opt}
Susan Zhang, Stephen Roller, Naman Goyal, Mikel Artetxe, Moya Chen, Shuohui
  Chen, Christopher Dewan, Mona Diab, Xian Li, Xi~Victoria Lin, et~al. 2022.
\newblock Opt: Open pre-trained transformer language models.
\newblock \emph{arXiv preprint arXiv:2205.01068}.

\end{thebibliography}
\bibliographystyle{acl_natbib}

\clearpage
\appendix

\label{sec:appendix}
\section{Data Preparation}
\label{Appendix: data Preparation}
\subsection{Preprocess of sentence-level corpora}
\paragraph{OpenI} In the original OpenI release, many non-sensitive terms were incorrectly masked as ``xxxx'' by the de-identification software described in~\cite{demner2016openi}. Our medical team manully fills in the missing information based on the context of the reports and additional information associated with it.
\subsection{Preprocess of the document-level corpra}
\paragraph{Manual De-identification Criteria} We hire human reviewers to manually inspect the reports after the automatic de-identification tools. According to our criteria, we will discard a datapoint if it contains
\begin{itemize}
    \item real names of the patient, or the healthcare professions,
    \item home address, working address, or locations of the patient or healthcare professionals.
    \item contact information (e.g., phone number) about the patient, or healthcare professionals. 
\end{itemize}

In the second round of human inspection about de-identification, 99.8\% of the data are well de-identified in the automatic stage, and 0.2\% of the data are discarded.

\paragraph{Diseases Code Preparation}
Before experts examine the disease codes of a report, we first group data by their resources. For reports sourced from MIMIC-CXR, we employ CheXpert \cite{irvin2019chexpert}, a rule-based automatic labeler to generate pseudo-labels for the diagnostic codes. Each label has three options: positive, negative, and unknown. In the case of reports from the second source, we customized a rule-based automatic labeler called \texttt{pyConText NLP}\footnote{\url{https://github.com/chapmanbe/pyConTextNLP}}, which generates pseudo-codes according to the keywords of each domain. In the last step, our medical team manually reviews, correct the codes where there is a conflict (e.g., positive ``no finding'' appears with certain positive diseases), and writes down the correct codes for each report.
\subsection{Medical Team}
\label{appendix: medical_team}
Our medical expert team consists of 4 members, including 2 senior board-certified radiologists with more than 15 years of experience in healthcare and a doctor who has more than 10 years of experience serving as a PI of medical research. We follow standard labeling practices, involving multiple rounds of iterative review by different experts until Cohen's kappa coefficient reaches 0.85. Any remaining disagreements are collectively resolved.

\section{Implementation details}
\paragraph{Models}
All adapted transformers are implemented based on the HuggingFace\footnote{\url{https://huggingface.co/models}} libraries. 
All prompted LLMs are implemented with its original release in their official webpage or GitHub. 

The model sizes are listed in the \cref{sec:related work}. 
\paragraph{Experiment Hyperparameters}
The configurations of querying LLMs are listed as follows:
\begin{table}[h]
\centering
\resizebox{\linewidth}{!}{%
\begin{tabular}{@{}lcc@{}}
\toprule
             &          GPT-3        & ChatGPT       \\ \midrule
Engine            & text-davinci-003 & gpt-3.5-turbo \\
Temperature       & 0                & 0             \\
Max Tokens        & 200              & 200           \\
Return N          & 1                & 1             \\
Top P             & 1                & 1             \\
Frequency Penalty & 0                & 0             \\
Presence Penalty  & 0                & 0             \\ \bottomrule
\end{tabular}%
}
\caption{Configuration of ChatGPT and GPT-3}
\label{tab:GPTconfiguration}
\end{table}

The configurations for fine-tuning adapted languages models are: learning rate=0.0001, weight decay=0, optimizer=Adam, training epoch=10, batch size=64, max length=256, 
All codes are implemented with Python3.8 and PyTorch1.7.1 with CUDA10.1. operated on Ubuntu (16.04.7 LTS) server with 2 NVIDIA GeForce GTS A6000 GPUs. Each has memory of 49GB.

\section{Prompts used for Querying LLM}
\label{Appendix: prompt}
\subsection{Number of examples in few-shot settings} In our study, we maintain a balanced and unbiased approach by setting the number of examples equal to the number of classes when prompting the language models (especially in NLU tasks). Additionally, we explore alternative numbers of examples, such as 1, 3, 5, 7, or 9. The NLU experiment results presented in our paper utilize a 2-shot approach, while the NLG results employ a 3-shot approach.

\subsection{Templates Used in Prompting}
{For zero-shot settings} we design prompts using the following template:

\texttt{
We are here to address a new task, [TASK NAME]. Given a {[LEVEL NAME]}, written by a radiologist, please  {[TASK CONTENT]}. Now here is a new {[LEVEL]}, tell me {[TASK CONTENT]}, without saying anything else.
Input: \{\}, Label:
}

{For few-shot settings} we insert examples extracted from the training set of each task into the prompts by:

\texttt{
We are here to address a new task, [TASK NAME]. Given a {[LEVEL NAME]}, written by a radiologist, please tell {[TASK CONTENT]}. 
Here are [N] examples. 
[EXAMPLE 1]...[EXAMPLE N]
Now here is a new [LEVEL], tell me [TASK CONTENT], without saying anything else.
Input: \{\}, Label:
}

The options for the placeholder are:
\begin{itemize}
    \item \texttt{TASK}: \{\textit{abnormal identification}, \textit{ambiguous identification, rewrite an ambiguous sentence to be less ambiguous, summarization }\}
    \item  \texttt{LEVEL NMAE}: \{\textit{sentence, report}\}.
    \item \texttt{TASK CONTENT}:\{\textit{Tell if the sentence indicates abnormal findings or not, Tell if it is ambiguous, a sentence is 
    defined to be ambiguous because of (1) 
    medical jargon with meanings different 
    from everyday general usage, such as 
    unremarkable; (2) contradictory findings in the same sentence; (3) misleading grammatical errors such as no period between full sentences, 
    , Observed pathology findings, The summary} \}
    \item \texttt{N}: \{1, 2, 3, 4, 5, 7, 9\}
\end{itemize}
In the following sections, we provide some examples of the prompts we use to query LLM. 
\subsection{Sentence-Level Tasks}
For sentence-level tasks, there are the following different types of prompts for NLU tasks: 

\paragraph{Zero-shot Classification for Abnormality} 

\texttt{We are here to address a new task, abnormal identification. Given a sentence from a radiology report, written by a radiologist, please tell if this sentence indicates abnormal findings or not. \\
Now here is a new sentence, tell me if it is normal or abnormal, without saying anything else. Sentence: \{\} Label:} 

\paragraph{Two-shot classification for Abnormality:}  

\texttt{We are here to address a new task, abnormal identification. Given a sentence from a radiology report, written by a radiologist, please tell if this sentence indicates abnormal findings or not. Here are 2 examples. \\ 
        Sentence: there is likely left basilar opacity. Label: abnormal. \\ 
        Sentence:  unchanged exam without acute abnormality. Label: normal. \\ 
        Now here is a new sentence, tell me if it is normal or abnormal, without saying anything else. \\
        Sentence: \{\} Label:
}

\paragraph{Zero-shot Classification for Ambiguity:} \texttt{Here is a task to classify ambiguous sentences. Given a sentence in a radiology report, written by a radiologist, please tell if it is ambiguous. A sentence is defined to be ambiguous because of (1) medical jargon with meanings different from everyday general usage, such as unremarkable; (2) contradictory findings in the same sentence; (3) misleading grammatical errors such as no period between full sentences. \\ Now given a new sentence, answer me with “ambiguous” or “unambiguous”, without saying anything else. Sentence: \{\} Label:}

\paragraph{Two-shot Classification for Ambiguity:} \texttt{Here is a task to classify ambiguous sentences. Given a sentence in a radiology report, written by a radiologist, please tell if it is ambiguous. A sentence is defined to be ambiguous because of (1) medical jargon with meanings different from everyday general usage, such as unremarkable; (2) contradictory findings in the same sentence; (3) misleading grammatical errors such as no period between full sentences. Here are 2 examples. \\
        Sentence: lungs are unremarkable. Label: ambiguous. \\ 
        Sentence: unchanged chronic appearance of the left lung. Label: unambiguous. \\
        Now given a new sentence, answer me with “ambiguous” or “unambiguous”, without saying anything else. \\
        Sentence: \{\} Label:
}

Similarly, we will add more examples for the prompts built for more than two-shot situations. 
For NLG tasks, we have the following prompts:

\paragraph{Zero-shot Rewriting tasks}
\texttt{Here is a task to rewrite ambiguous sentences to be less ambiguous. Given a sentence in a radiology report, written by a radiologist, please rewrite it to be more explicit about the diagnostic decision reflected in the sentence, however, maintain the main meaning of the original sentence.  A sentence is defined to be ambiguous because of (1) medical jargon with meanings different from everyday general usage, such as unremarkable; (2) contradictory findings in the same sentence; (3) misleading grammatical errors such as no period between full sentences.  Now given a new sentence, answer me with its rewrite, without saying anything else. Sentence: \{\}. Rewrite:}

\paragraph{Three-shot Rewriting tasks}
\texttt{Here is a task to rewrite ambiguous sentences to be less ambiguous. Given a sentence in a radiology report, written by a radiologist, please rewrite it to be more explicit about the diagnostic decision reflected in the sentence, however, maintain the main meaning of the original sentence.  A sentence is defined to be ambiguous because of (1) medical jargon with meanings different from everyday general usage, such as unremarkable; (2) contradictory findings in the same sentence; (3) misleading grammatical errors such as no period between full sentences. Here are three examples.\\ Sentence: Lungs are unremarkable. Diagnostic: Normal. Rewrite: Lungs are normal.\\ Sentence: The lung volumes are low normal. Diagnostic: Normal. Rewrite: The lung volumes are in the lower range of normal limit. \\ Sentence: Cardiomegaly and hiatal hernia without an acute abnormality identifie. Diagnostic: Abnormal. Rewrite: Cardiomegaly and hiatal hernia . Without an acute abnormality identified. \\ Now given a new sentence, answer me with its rewrite, without saying anything else. Sentence: \{\}. Rewrite:}

\subsection{Document-level Tasks}
For NLU tasks, we use the following prompts:
\paragraph{Zero-shot Summarization Tasks}
\texttt{Here is a summarization task. Given a radiology report written by a radiologist, please write a summary of the report. Answer me with its summary only, without saying anything else. Report: \{\}. Summary:}

\paragraph{Three-shot Summarization Tasks}
\texttt{Here is a summarization task. Given a radiology report written by a radiologist, please write a summary of the report. Here are 5 examples.\\
        Report: clinical history 62-year-old male with bilateral bunions. please perform weighted views. a lateral weight bearing view of bilateral feet as well as a lateral non-weight bearing view of bilateral feet are studied. three views of the right foot show no fracture dislocation foreign body pathologic calcification or soft tissue swelling. the joint spaces are not noticeable. there is a deformity of the hallux valgus. the angle of pitch is within normal limits. three views of the left foot show no fracture dislocation foreign body pathologic calcification or soft tissue swelling. the joint spaces are not noticeable. hallux valgus minor. the angle of pitch is within normal limits. Summary: 1 bilateral hallux valgus deformities. 2 no acute osseous abnormalities. ssn7312ptc1job no. 1157. \\
        Report: there has been no significant change in the patient's condition since the patient's exam which was earlier in the day. the heart size is normal and the lungs are free of disease. on the left base is again noted a small granulom.  Summary: for a active disease in the chest there is no evidence. \\
        Report: the cardiovascular-mediastinal silhouette is normal. it's not unusual for pulmonary vessels. the bones appear to be intact. Summary: chest x-rays within normal ranges. no change in date 2010-06-28.\\
        Now given a new report, answer me with its summary only, without saying anything else. Report: \{\}. Summary: }

\section{More Discussions}
\subsection{Case Studies of BioMed LM}
\label{appendx:biomed}
BioMed LM, a GPT-style LLM trained on PubMed, lacks instruction fine-tuning in its training process. As a result, the model's outputs often lack a unified format, making it challenging to conduct follow-up statistical evaluations. To assess the model's performance, we introduce the concept of a qualified rate, which represents the proportion of test cases where the model provides a relevant prediction for the given task, such as identifying abnormalities by including only one of the ``normal'' or ``abnormal'' in the response. We report the qualified rate here:
\begin{table}[h]
\centering
\resizebox{0.7\linewidth}{!}{%
\begin{tabular}{@{}ccc@{}}
\toprule
              & Sentence & Report \\ \midrule
Zero-shot NLU & 40\%     & 1\%    \\
Few-shot NLU  & 92\%     & 85\%   \\ 
\bottomrule
\end{tabular}%
}
\caption{Qualified Rate of BioMed LM on NLU tasks}
\label{tab:qr biomed LM}
\end{table}

In zero-shot settings, BioMed LM struggles to adhere to the instructions in the prompt, generating outputs that are freestyle and not aligned with the expected format. In a few-shot setting, particularly in document-level tasks, the length of the prompts exceeds the maximum input capacity of BioMed LM (1024 tokens). To address this issue, we employ two solutions: (1) chunking the input into 1024-token segments and (2) discarding test cases that exceed the maximum length. However, both solutions have drawbacks. The chunked uncompleted input approach leads to more   ``freestyle'' outputs, resulting in a lower qualified rate. On the other hand, discarding examples introduces a high variance in the statistics. We show some examples of the ``freestyle'' outputs at the end of this section. 

Considering these challenges, we treat BioMed LM as an exception and exclude it when analyzing results across different categories of LMs. Nonetheless, we emphasize the importance of further investigation into biomedical LLMs for future research.

\paragraph{Example Output of BioMed LM}

We show some examples of unqualified ``free-style'' outputs from BioMed LM in the following:\\ 
\paragraph{Sentence Level, NLU-Zero-shot}
Note that we expect the model to return a single word from ``normal'', ``abnormal'', ``ambiguous'', or ``unambiguous''. However, BioMed LM returns the following output:

\noindent Prompt: \texttt{Given a sentence from a radiology report, written by a radiologist, please tell if the following sentence indicates "normal" or "abnormal" findings. Sentence: median sternotomy wires appear intact. Label:} 

\noindent Returned Text: \texttt{median sternotomy wires appear intact.}

\paragraph{Sentence Level, NLU-Two-shot} 
In this case, we expect the model to return a single word ``normal'', ``abnormal'', ``ambiguous'', ``unambiguous''. However, the output from BioMed is:

\noindent Prompt: \texttt{Here is a task to classify ambiguous sentences. Given a sentence in a radiology report, written by a radiologist, please tell if it is ambiguous. A sentence is defined to be ambiguous because of (1) medical jargon with meanings different from everyday general usage, such as unremarkable; (2) contradictory findings in the same sentence; (3) misleading grammatical errors such as no period between full sentences. Here are 2 examples. \\
Sentence: lungs are unremarkable. Label: ambiguous.\\ 
Sentence: unchanged chronic appearance of the left lung. Label: unambiguous. 
\\
Now given a new sentence, answer me with “ambiguous” or “unambiguous”.\\
Sentence: findings: there is a cystic, approximately 3. 6 x 2. 
6 x 6. 6 cm, mass with its epicenter in the left submandibular region, with extension into the left submental space, left sublingual space, left base of tongue, and superiorly into the left peripharyngeal space. Label:}

\noindent Returned Text: \texttt{cystic mass. \\
Sentence: there is a cystic, approximately 3. 6 x 2. 6 x 6. 6 cm, mass with its epicenter in the left submandibular region, with extension into the}

\paragraph{Sentence Level, NLG-Zero-shot}
Here we show an unqualified example in the ambiguous rewriting task, where the output does not follow the instruction of ``without saying anything else''. 

Prompt: \texttt{Here is a task to rewrite ambiguous sentences to be less ambiguous. Given a sentence in a radiology report, written by a radiologist, please rewrite it to be more explicit about the diagnostic decision reflected in the sentence, however, maintain the main meaning of the original sentence.  A sentence is defined to be ambiguous because of (1) medical jargon with meanings different from everyday general usage, such as unremarkable; (2) contradictory findings in the same sentence; (3) misleading grammatical errors such as no period between full sentences.  Now given a new sentence, answer me with its rewrite, without saying anything else. Sentence: aorta appears unchanged as compared to the study. Diagnostic: Abnormal. Rewrite: }

\noindent Returned Text: \texttt{$\backslash\backslash$... aorta appears unchanged as compared to the study.\\-----------------------------------------------------------------------------------------------------------------------}

In this unqualified example, the BioMed LM does not follow the instruction and just output an irrelevant sentence.

\noindent Prompt: \texttt{Here is a task to rewrite ambiguous sentences to be less ambiguous. Given a sentence in a radiology report, written by a radiologist, please rewrite it to be more explicit about the diagnostic decision reflected in the sentence, however, maintain the main meaning of the original sentence.  A sentence is defined to be ambiguous because of (1) medical jargon with meanings different from everyday general usage, such as unremarkable; (2) contradictory findings in the same sentence; (3) misleading grammatical errors such as no period between full sentences.  Now given a new sentence, answer me with its rewrite, without saying anything else. Sentence: apparent scarring within the lingula otherwise unremarkable. Diagnostic: Abnormal. Rewrite: }

\noindent Returned Text:
\texttt{$\backslash\backslash$[Lingula$\backslash\backslash$] scarring.\\-----------------------------------------------------------------------------------------------------------------------}

\paragraph{Document Level, NLU-Zero-shot}
In this task, we expect the output to have a prediction for every disease. However, BioMed LM just repeats the disease code as the output. 

Prompt: \texttt{Here is a task to classify the observations of the chest reports. Given a radiology report written by a radiologist, please predict if every one of the following 14 observations is positive, negative, or unknown. The 14 observations are No Finding, Enlarged Cardiomediastinum, Cardiomegaly, Lung Lesion, Lung Opacity, Edema, Consolidation, Pneumonia, Atelectasis, Pneumothorax, Pleural Effusion, Pleural Other, Fracture, Support Devices. Answer me only with the tuples of each observation and its preidction, without saying anything else. There should be 14 tuples in your answer and tuples are separated by comma. Report: bibasilar changes in atelectatics. the lungs are otherwise clear. the cardiomedical silhouette and the hila are normal. right port-a-cath ends at the cavoatrial junction. there's no pneumothorax in this case. No  pleural effusion. Mentioned observations:}

\noindent Returned Text: \texttt{No finding, Enlarged Cardiomediastinum, Cardiomegaly, Lung Lesion, Lung Opacity, Edema, Consolidation, Pneumonia, Atelectasis, Pneumothorax, Pleural Effusion, Pleural Other}.

\paragraph{Document-Level, NLG-Two-shot}
We show an example of ``freestyle'' output from BioMed LM

Prompt: \texttt{Here is a summarization task. Given a radiology report written by a radiologist, please write a summary of the report. Here are 2 examples.\\Report: clinical history 62-year-old male with bilateral bunions . please perform weighted views. a lateral weight bearing view of bilateral feet as well as a lateral non-weight bearing view of bilateral feet are studied. three views of the right foot show no fracture dislocation foreign body pathologic calcification or soft tissue swelling. the joint spaces are not noticeable. there is a deformity of the hallux valgus. the angle of pitch is within normal limits. three views of the left foot show no fracture dislocation foreign body pathologic calcification or soft tissue swelling. the joint spaces are not noticeable. hallux valgus minor. the angle of pitch is within normal limits. Summary: 1 bilateral hallux valgus deformities. 2 no acute osseous abnormalities. ssn7312ptc1job no. 1157. \\Report: there has been no significant change in the patient's condition since the patient's exam which was earlier in the day. the heart size is normal and the lungs are free of disease. on the left base is again noted a small granulom.  Summary: for a active disease in the chest there is no evidence. \\Report: right and left feet have severe hammertoes. calcaneus on the left shows an achilles spur. the severe hallux valgus configuration of the feet is observed. a dominant finding was the degree of luxation and deviation of first metatarsal proximal phalangeal joint.. Summary:}

\noindent Returned Text: \texttt{1 bilateral hammertoes. 2 no acute osseous abnormalities. ssn7312ptc1job no. 1157.\\Report: the patient has been seen in the clinic for a routine checkup. The patient has}

\section{More Results}
\label{appendix: more_results}
We considered popular LLMs such as GPT3, ChatGPT, and Vicuna-7B (the instruction finetuned LLaMa) in the main context. But we also keep in mind that LLMs are rapidly developing and we are following up by adding the newest models into our evaluation. In \cref{tab:more_results_NLU}, \cref{tab: more_results_NLG},  \cref{tab:more_results_document_NLU}, and \cref{tab:more_results_document_NLG},  we provide more results with GPT4\footnote{\url{https://openai.com/gpt-4}}, LLaMa2 \cite{touvron2023llama}, LLaMa2-chat \cite{touvron2023llama}, GPT-NeoX\ cite{gpt-neox-20b}, PMC-LLaMa \cite{wu2023pmcllama}, BioGPT \cite{10.1093/bib/bbac409} etc, which align with the findings already in the paper. We will keep working on follow-ups and add more evaluation results here (and also in our Github repository):

\begin{table*}[t]
\centering
\resizebox{0.85\textwidth}{!}{%
\begin{tabular}{@{}ccllll@{}}
\toprule
 &
  \multirow{2}{*}{Models} &
  \multicolumn{2}{c}{Chest} &
  \multicolumn{2}{c}{Miscellaneous Domains} \\ \cmidrule(l){3-6} 
 &
   &
  \multicolumn{1}{c}{Abnormality$\uparrow$ } &
  \multicolumn{1}{c}{Ambiguity$\uparrow$ } &
  \multicolumn{1}{c}{Abnormality$\uparrow$ } &
  \multicolumn{1}{c}{Ambiguity$\uparrow$ } \\ \midrule
\multirow{4}{*}{\begin{tabular}[c]{@{}c@{}}Adapted PLMs with\\ Fine-Tuning\end{tabular}} &
  zero-shot BioGPT &
  0.7521 &
  0.3012 &
  0.6994 &
  0.0543 \\
 & few-shot BioGPT       & 0.5966 & 0.2864 & 0.5990 & 0.0672 \\
 & zero-shot PMC-LLaMa   & 0.6663 & 0.2986 & 0.6189 & 0.0465 \\
 & few-shot PMC-LLaMa    & 0.7660 & 0.2606 & 0.7162 & 0.0574 \\ \midrule
\multirow{8}{*}{\begin{tabular}[c]{@{}c@{}}LLMs Prompted by\\ Zero/Few Shot\end{tabular}} &
  zero-shot GPT4 &
  0.9299 &
  0.6596 &
  0.9547 &
  0.8462 \\
 & few-shot GPT4         & 0.9145 & 0.7925 & 0.9417 & 0.8906 \\
 & zero-shot GPT-NeoX    & 0.6371 & 0.2747 & 0.5460 & 0.0651 \\
 & few-shot GPT-NeoX     & 0.5986 & 0.2578 & 0.5000 & 0.0648 \\
 & zero-shot LLaMa2-chat & 0.4626 & 0.2864 & 0.4075 & 0.0679 \\
 & few-shot LLaMa2-chat  & 0.6551 & 0.2859 & 0.5685 & 0.0679 \\
 & zero-shot LLaMa2      & 0.5568 &	0.3150	& 0.6474	& 0.0679 \\
 & few-shot LLaMa2       & 0.4728 & 0.2864 & 0.4240 & 0.0679 \\ \bottomrule
\end{tabular}%
}
\caption{More results on sentence-level NLU tasks. }
\label{tab:more_results_NLU}
\end{table*}

\begin{table*}[ht]
\centering
\resizebox{\textwidth}{!}{%
\begin{tabular}{@{}ccllllll@{}}
\toprule
 &
  \multirow{2}{*}{Models} &
  \multicolumn{3}{c}{Chest} &
  \multicolumn{3}{c}{Miscellaneous Domains} \\ \cmidrule(l){3-8} 
 &
   &
  \multicolumn{1}{c}{\begin{tabular}[c]{@{}c@{}}Disambiguation\\ $\Delta Acc_{am} \uparrow$\end{tabular}} &
  \multicolumn{1}{c}{\begin{tabular}[c]{@{}c@{}}Content Distortion\\ $\Delta Acc_{ab} \downarrow$\end{tabular}} &
  BLEU4 $\uparrow$ &
  \multicolumn{1}{c}{\begin{tabular}[c]{@{}c@{}}Disambiguation\\ $\Delta Acc_{am} \uparrow$\end{tabular}} &
  \multicolumn{1}{c}{\begin{tabular}[c]{@{}c@{}}Content Distortion\\ $\Delta Acc_{ab} \downarrow$\end{tabular}} &
  BLEU4 $\uparrow$ \\ \midrule
\multirow{4}{*}{\begin{tabular}[c]{@{}c@{}}Adapted PLMs with\\ Fine-Tuning\end{tabular}} &
  zero-shot BioGPT &
  0.8184 &
  0.2971 &
  19.54 &
  0.8623 &
  0.4179 &
  19.03 \\
 &
  few-shot BioGPT &
  0.8019 &
  0.3036 &
  22.43 &
  0.8651 &
  0.4382 &
  22.32 \\
 &
  zero-shot PMC-LLaMa &
  0.8119 &
  0.4389 &
  19.03 &
  0.9235 &
  0.5393 &
  18.77 \\
 &
  few-shot PMC-LLaMa &
  0.5577 &
  0.2244 &
  22.61 &
  0.6719 &
  0.3213 &
  21.07 \\ \midrule
\multirow{8}{*}{\begin{tabular}[c]{@{}c@{}}LLMs Prompted\\ by\\ Zero/Few Shot\end{tabular}} &
  zero-shot GPT4 &
  0.7921 &
  -0.0297 &
  60.61 &
  0.7562 &
  0.1652 &
  60.36 \\
 &
  few-shot GPT4 &
  0.7591 &
  0.0000 &
  63.66 &
  0.7753 &
  0.0697 &
  67.67 \\
 &
  zero-shot GPT-NeoX &
  0.4125 &
  0.0462 &
  38.23 &
  0.4786 &
  0.2225 &
  35.16 \\
 &
  few-shot GPT-NeoX &
  -0.056 &
  0.0396 &
  24.65 &
  0.0876 &
  0.2517 &
  25.91 \\
 &
  zero-shot LLaMa2-chat &
  0.4059 &
  0.0033 &
  53.23 &
  0.3379 &
  0.1314 &
  53.20 \\
 &
  few-shot LLaMa2-chat &
  0.4521 &
  0.0198 &
  57.26 &
  0.4480 &
  0.1066 &
  52.59 \\
 &
  zero-shot LLaMa2 &
  0.1320 &
  0.0165 &
  29.95 &
  0.0921 &
  0.0157 &
  39.60 \\
 &
  few-shot LLaMa2 &
  0.1386 &
  0.257 &
  10.77 &
  0.4044 &
  0.1101 &
  10.18 \\ \bottomrule
\end{tabular}%
}
\caption{More results on sentence-level NLG tasks. }
\label{tab: more_results_NLG}
\end{table*}

\begin{table*}[ht]
\centering
\resizebox{\textwidth}{!}{%
\begin{tabular}{@{}cccccccc@{}}
\toprule
 &
  \multirow{2}{*}{Models} &
  \multicolumn{2}{c}{Chest} &
  \multicolumn{2}{c}{Foot} &
  \multicolumn{2}{c}{Ankle} \\ \cmidrule(l){3-8} 
 &                       & avg Accuracy$\uparrow$  & avg EMR$\uparrow$  & avg Accuracy$\uparrow$  & avg EMR$\uparrow$  & avg Accuracy$\uparrow$  & avg EMR$\uparrow$  \\ \midrule
\multirow{4}{*}{\begin{tabular}[c]{@{}c@{}}Adapted PLMs with\\ Fine-Tuning\end{tabular}} &
  zero-shot BioGPT &
  0.8276 &
  0.0654 &
  0.9222 &
  0.0010 &
  0.9299 &
  0.0020 \\
 & few-shot BioGPT       & 0.7561       & 0.0030  & 0.9618       & 0.4196  & 0.9688       & 0.4860  \\
 & zero-shot PMC-LLaMa   & 0.8290       & 0.0663  & 0.9228       & 0.0010  & 0.9206       & 0.0010  \\
 & few-shot PMC-LLaMa    & 0.7549       & 0.0039  & 0.9604       & 0.4044  & 0.9228       & 0.0020  \\ \midrule
\multirow{6}{*}{\begin{tabular}[c]{@{}c@{}}LLMs Prompted\\ by\\ Zero/Few Shots\end{tabular}} &
  zero-shot GPT-NeoX &
  0.8291 &
  0.0663 &
  0.9238 &
  0.0033 &
  0.9318 &
  0.0194 \\
 & few-shot GPT-NeoX     & 0.7876       & 0.0537  & 0.9617       & 0.4186  & 0.9691       & 0.4908  \\
 & zero-shot LLaMa2-chat & 0.3192       & 0.0294  & 0.9617       & 0.4186  & 0.9691       & 0.4908  \\
 & few-shot LLaMa2-chat  & 0.8192       & 0.0505  & 0.9617       & 0.4186  & 0.9691       & 0.4908  \\
 & zero-shot LLaMa2      & 0.7995       & 0.0591  & 0.9518       & 0.3171  & 0.9669       & 0.4597  \\
 & few-shot LLaMa2       & 0.8331       & 0.0722  & 0.9594       & 0.3898  & 0.9478       & 0.1135  \\ \bottomrule
\end{tabular}%
}
\caption{More results on document-level NLU tasks. }
\label{tab:more_results_document_NLU}
\end{table*}

\begin{table*}[t]
\centering
\resizebox{0.85\textwidth}{!}{%
\begin{tabular}{@{}ccccccc@{}}
\toprule
 & \multirow{2}{*}{Models} & \multicolumn{5}{c}{Miscellaneous Domains}      \\ \cmidrule(l){3-7} 
 &                         & ROUGE-1$\uparrow$  & ROUGE-2$\uparrow$  & ROUGE-L$\uparrow$  & Sum $\uparrow$      & BLEU4 $\uparrow$ \\ \midrule
\multirow{4}{*}{\begin{tabular}[c]{@{}c@{}}Adapted PLMs with\\ Fine-Tuning\end{tabular}}     & zero-shot BioGPT & 18.1577 & 6.5828  & 16.9262 & 41.6667  & 8.10  \\
 & few-shot BioGPT         & 12.3569 & 3.8041  & 11.6335 & 27.7945  & 5.26  \\
 & zero-shot PMC-LLaMa     & 11.6813 & 3.8402  & 10.8875 & 26.409   & 7.52  \\
 & few-shot PMC-LLaMa      & 14.5339 & 3.9377  & 13.0060 & 31.4776  & 8.46  \\ \midrule
\multirow{9}{*}{\begin{tabular}[c]{@{}c@{}}LLMs Prompted\\ by\\ Zero/Few Shots\end{tabular}} & zero-shot GPT4   & 59.8081 & 34.6459 & 55.1671 & 149.6211 & 53.82 \\
 & few-shot GPT4           & 56.1813 & 31.6360 & 51.5054 & 139.3228 & 44.16 \\
 & zero-shot GPT-NeoX      & 18.3383 & 6.6764  & 17.0308 & 42.0455  & 11.01 \\
 & few-shot GPT-NeoX       & 12.3496 & 3.8196  & 11.6543 & 27.8235  & 8.11  \\
 & zero-shot LLaMa2-chat   & 21.0446 & 6.9982  & 18.8494 & 46.8923  & 21.97 \\
 & few-shot LLaMa2-chat    & 17.8764 & 5.6633  & 15.5410 & 39.0806  & 21.54 \\
 & zero-shot LLaMa2        & 18.0240 & 5.8142  & 16.5904 & 40.3801  & 15.37 \\
 & few-shot LLaMa2         & 21.4236 & 7.8157  & 20.0528 & 49.2921  & 18.92 \\ \bottomrule
\end{tabular}%
}
\caption{More results on document-level NLG tasks. }
\label{tab:more_results_document_NLG}
\end{table*}

\end{document}